\documentclass[10pt,letterpaper]{article}
\usepackage[top=0.85in,left=2.75in,footskip=0.75in,marginparwidth=2in]{geometry}

\usepackage[utf8]{inputenc}

\usepackage[numbers]{natbib}
\usepackage{nameref,hyperref}

\usepackage[right]{lineno}

\usepackage{microtype}
\DisableLigatures[f]{encoding = *, family = * }

\raggedright
\usepackage{changepage}

\usepackage[aboveskip=1pt,labelfont=bf,labelsep=period,singlelinecheck=off]{caption}

\makeatletter
\renewcommand{\@biblabel}[1]{\quad#1.}
\makeatother

\usepackage{lastpage,fancyhdr,graphicx}
\usepackage{epstopdf}
\fancyheadoffset[L]{2.25in}
\fancyfootoffset[L]{2.25in}

\usepackage{color}

\definecolor{Gray}{gray}{.25}

\usepackage{graphicx}

\usepackage{sidecap}

\usepackage{wrapfig}
\usepackage[pscoord]{eso-pic}
\usepackage[fulladjust]{marginnote}
\reversemarginpar

\usepackage{amssymb}
\usepackage{booktabs}
\usepackage{algorithm}
\usepackage[noend]{algpseudocode}
\usepackage{bm}
\usepackage{svg}
\usepackage{gensymb}
\usepackage{adjustbox}
\usepackage{tikz}
\usepackage{multirow}
\usepackage{enumitem}
\usepackage{tcolorbox}
\usetikzlibrary{decorations.pathreplacing,calc}
\usepackage{optidef}
\usepackage{algorithm}
\usepackage{algpseudocode}
\usepackage{comment}
\usepackage{subcaption}
\usepackage{setspace}
\usetikzlibrary{shapes.geometric, arrows.meta, positioning}
\usepackage{nomencl}
\usepackage{etoolbox}
\usetikzlibrary{shapes.geometric, arrows}

\usepackage{amsmath}
\usepackage{cleveref}
\usepackage{siunitx}
\usepackage{tikz}

\providecommand{\ac}[1]{#1}

\providecommand{\acp}[1]{#1s}

\begin{document}
\vspace*{0.35in}

\begin{flushleft}
{\Large
\textbf\newline{An Autonomous, End-to-End, Convex-Based Framework for Close-Range Rendezvous Trajectory Design and Guidance with Hardware Testbed Validation}
}
\newline
Minduli C. Wijayatunga$^{2}$, Julian Guinane$^{1}$, Nathan D. Wallace$^{1}$, Xiaofeng Wu$^{1}$
\\
$^{1}$ Massachusetts Institute of Technology, Cambridge, MA 02139, United States \\
$^{2}$ University of Sydney, Chippendale, New South Wales, NSW 2008, Australia
\end{flushleft}

\section*{Abstract}
Autonomous satellite servicing missions must execute close-range rendezvous under stringent operational and safety constraints, all while dealing with uncertainties in sensing, actuation, and dynamics, while remaining computationally tractable for onboard implementation. This paper presents CORTEX (Convex Optimization for Rendezvous Trajectory Execution), an autonomous, perception-enabled, real-time trajectory design and guidance framework for close-range rendezvous. CORTEX combines a deep-learning-enabled perception pipeline with a convex-optimization-based trajectory design and guidance framework, which also includes reference regeneration and abort-to-safe-orbit capabilities to account for major trajectory deviations resulting from unexpected sensor errors and engine failure events. 
CORTEX has been validated under high-fidelity software simulations and hardware-in-the-loop experiments. The software pipeline was developed using Basilisk and incorporates high-fidelity relative dynamics, realistic thruster execution, perception, and attitude control.
Hardware-in-the-loop testing comprised of an optical navigation testbed to evaluate perception-to-estimation performance, and a planar air-bearing testbed to evaluate the end-to-end trajectory design and guidance loop under representative actuation and subsystem effects. 
A Monte-Carlo campaign was conducted in the software testbed, which incorporated initial state uncertainty, thrust magnitude errors, and missed-thrust events. Under the strongest error case investigated, CORTEX achieves terminal docking performance of $36.85 \pm 44.46$~mm in relative position and $1.25 \pm 2.26$~mm/s in relative velocity. 
18 test cases were selected for execution on the planar air-bearing testbed with optitrack, comprising 10 nominal cases not requiring reference recomputation or abort, and 8 off-nominal cases in which constraint violations were produced by simulating engine failure and unexpected sensor malfunctions, requiring recomputations and abort-to-safe-orbit operations.  The resulting terminal docking errors for these experiments were $8.09 \pm 5.29$ mm in position and $2.23 \pm 1.72$ mm/s in velocity. 
\section*{Keywords}
Convex Guidance; Close-range Rendezvous; Hardware-in-the-loop testing; Basilisk

\section{Introduction}

\ac{RPO} are shifting from bespoke demonstration missions such as  Northrop Grumman’s Mission Extension Vehicles \cite{pyrak2022performance}, China’s Tianyuan--1 \cite{davis2019orbit}, and Astroscale's ADRAS-J \cite{lidtke2025telescope},
to routine infrastructure for modern spaceflight, enabled by the growth of in-orbit assets and the resulting demand for inspection, servicing, assembly, and debris mitigation.
The number of tracked objects in low Earth orbit has grown from about 13,700 in 2019 to 24,185 in 2025, an increase of roughly 76\% in six years \cite{Pultarova2025_leo_crowded}. 
This growth implies that RPO must scale in cadence and cost, making autonomy a prerequisite for economically viable and repeatable on-orbit operations.

Among \ac{RPO} phases, the close-range rendezvous, where the servicer locates the docking port axes and 
approaches the client, is particularly safety-critical.
It must satisfy strict operational constraints (e.g., keep-out zones, approach corridors, plume/line-of-sight constraints, illumination/eclipses) under limited actuation authority and imperfect relative navigation \cite{fehse2003automated}.
In this phase, spacecraft must not only generate and track trajectories autonomously but also make operational go/no-go decisions and execute safe contingency actions, such as hold, retreat, and replan, when state or thrust uncertainties exceed expected bounds. Deployable autonomy for close-proximity flight must also be reliable, computationally predictable for onboard execution, and amenable to verification and validation.
These requirements place a premium on methods with explicit constraint handling and analyzable behaviour, rather than solely empirical performance.

\ac{RPO} trajectory design typically involves determining the burn times and thrust vectors while optimizing mission objectives such as propellant consumption, time of flight, inspection quality, safety, and robustness \cite{BORELLI202475}. This process must also satisfy a variety of constraints, including sun-illumination windows, terminal position and velocity conditions, minimum burn intervals, navigation sensor limitations, and passive safety requirements. These methods are commonly categorized into three classes: linear rendezvous, nonlinear two-body rendezvous, and perturbed or constrained rendezvous \cite{LUO20141}. Linear approaches generally rely on the \ac{CW} equations, and include both analytical and numerical techniques. Analytical strategies are often derived from Lawden’s primer vector theory \cite{Prussing}, which are effective when optimizing only propellant or manoeuvre direction. Numerical approaches include genetic algorithms, indirect methods, pseudospectral methods and convex optimization \cite{BORELLI202475, geller2017real}. Among these, convex optimization stands out for its ability to provide real-time performance, enabling onboard replanning during dynamic mission phases \cite{geller2017real}. While convex methods have been applied to trajectory design in literature, to our knowledge they have rarely been tested in fully integrated simulation environments.  Nonlinear two-body rendezvous methods are typically used during earlier mission phases, when the chaser-target separation is large. These methods often involve solving multi-revolution Lambert problems \cite{torre2015review} and applying nonlinear optimization to account for full two-body dynamics and perturbations \cite{WIJAYATUNGA2025109996}.

Guidance algorithms bridge these planning methods with the realities of imperfect actuation and navigation.  Guidance schemes for the close-range rendezvous must map estimated relative states to thrust commands that track a reference trajectory while compensating for thrust errors, modelling discrepancies, and measurement noise~\cite{doi:10.2514/1.G006191}. For autonomous close proximity, effective schemes must achieve this with low computational overhead, reliable onboard convergence, and explicit constraint handling.  Recent advancements in \ac{RPO} guidance have focused on methods that balance optimality and onboard feasibility, such as reinforcement learning \cite{WIJAYATUNGA2025109996}, behaviour cloning approaches \cite{doi:10.2514/1.A34838} and convex optimization frameworks \cite{doi:10.2514/1.G008179}. They offer varying trade-offs between computational complexity, adaptability, and adherence to constraints, with convex and learning-based approaches showing significant promise for real-time, constraint-aware guidance. Among these, convex approaches are particularly well-suited to verification and validation, as their behaviour can be analysed through the structure of the optimisation problem itself (e.g., constraint sets, terminal conditions, and abort modes). While convex-guidance schemes for \ac{RPO} are commonly present in literature, replanning and abort guidance strategies as well as the process of making go/no-go decisions are not well explored. 

This work proposes \ac{CORTEX}, a convex optimization-based framework for trajectory design and guidance during the close-range rendezvous phase.  Within \ac{CORTEX}, relative navigation is provided by a deep-learning perception pipeline based on \ac{YOLO} keypoint detection, integrated with an \ac{EKF} that incorporates new measurements when keypoints are visible, but otherwise propagates the relative state forward using the dynamics model. The trajectory design is carried out by a two-level convex optimiser that jointly enforces eclipse avoidance constraints, keep-out sphere constraints, plume-impingement avoidance, and approach corridor requirements. Guidance is implemented via a convex adaptive controller, and automatic reference regeneration and abort-to-safe-orbit capabilities are also introduced to account for major trajectory deviations due to unexpected sensor errors and engine failure events. 

The trajectory design component of \ac{CORTEX} builds on prior efforts such as \cite{geller2017real} and extends it to account for eclipse constraints while handling both the fly-around and docking phases in a unified framework. Instead of embedding eclipse constraints directly within a single large convex program,  \emph{sunlight-aware hold points} are introduced to impose explicit waiting times to avoid entering critical inspection and final-approach segments during eclipses, while each phase trajectory is still generated via convex optimisation under standard dynamical and safety constraints. This coupling between eclipse-aware phase scheduling and convex tracking guidance yields trajectories that are both fuel-efficient and safe, while adhering to realistic illumination constraints.

The \ac{CORTEX} guidance component draws inspiration from \cite{doi:10.2514/1.G008089}, which introduces a convex tracking approach for low-thrust transfers in active debris removal missions. In this work, this framework is extended to impulsive proximity operations. While the core architecture---comprising reference trajectory generation, convex tracking, and optional reference regeneration---is retained from \cite{doi:10.2514/1.G008089}, new reference generation/regeneration strategies are introduced in \ac{CORTEX}. Additionally, an abort-on-command capability is added to guarantee on-demand ending of the mission.

As the goal of this work is the development of space autonomy suitable for mission adoption, validation must extend beyond open-loop or simplified simulation. As such, a two-tier validation campaign is employed, which entails: 
\begin{enumerate}
    \item \textbf{A Software Simulation Testbed for Monte-Carlo Testing}: a Basilisk-based \cite{doi:10.2514/1.I010762} high fidelity simulation of spacecraft motion under high-order orbital perturbations, drag, and third-body effects within a closed-loop system that also considers perception. This is used for conducting Monte Carlo testing under various thrust and state error magnitudes and engine failure probabilities, to statistically validate the performance of \ac{CORTEX}.
    \item \textbf{A Planar Air bearing Testbed for End-to-End Testing}: An air-bearing table setup and test satellites with full spacecraft subsystems onboard, used to test 4 nominal test cases and 4 off-nominal test cases where engine failures and large sensor errors are emulated to demonstrate the use of recomputations and aborts in the mission.
\end{enumerate}

This structure enables the quantitative assessment of performance, robustness, and the frequency/character of contingency-triggered events.


In summary, the key contributions of this work include: 
\begin{enumerate}[itemsep=0.5ex]
    \item Hierarchical close-range reference generation that couples eclipse-aware phase scheduling (hold-point logic) with convex trajectory generation for fly-around and docking phases.
    \item Single-iteration, convex adaptive tracking guidance tailored to impulsive proximity operations, with explicit convex constraint handling and predictable per-cycle solve time.
    \item Operational autonomy logic for close-range RPO, including phase supervision, reference regeneration triggers, and abort-to-safe-orbit capability under off-nominal sensing/actuation conditions.
    \item Closed-loop validation in a high-fidelity Basilisk simulation including perturbations, thrust/state uncertainty, missed-thrust events, and (where applicable) perception-driven measurement updates.
    \item Hardware-in-the-loop demonstration on a planar air-bearing testbed, exercising nominal and off-nominal scenarios (replan/abort) in real time.
\end{enumerate}

The remainder of this paper is organised as follows:
Section~\ref{MissionDescription} describes the mission description and dynamical modelling involved,
Section~\ref{ssec:TrajDesg} presents the \ac{CORTEX} trajectory design and guidance methodology,
Section~\underline{[Z]} details \underline{[perception/estimation and integration]},
and Section~\underline{[W]} reports \underline{[simulation and experimental validation results]}.

\section{Mission Description and Modelling}\label{MissionDescription}


The \ac{RPO} mission considered in this study is an \ac{ISAM} mission, designed to demonstrate autonomous approach, servicing, and inspection capabilities for an on-orbit \ac{CSO} located in \ac{LEO}. The mission architecture features a Servicer equipped with a comprehensive \ac{GNC} system for proximity operations. 
A high-level overview of the mission architecture is presented in \cite{wallace2025iac} and \cite{guinane2025iac}.

%


\subsection{Mission Concept of Operations}

\begin{figure}[tbp]
    \centering
    \includegraphics[width=0.8\linewidth]{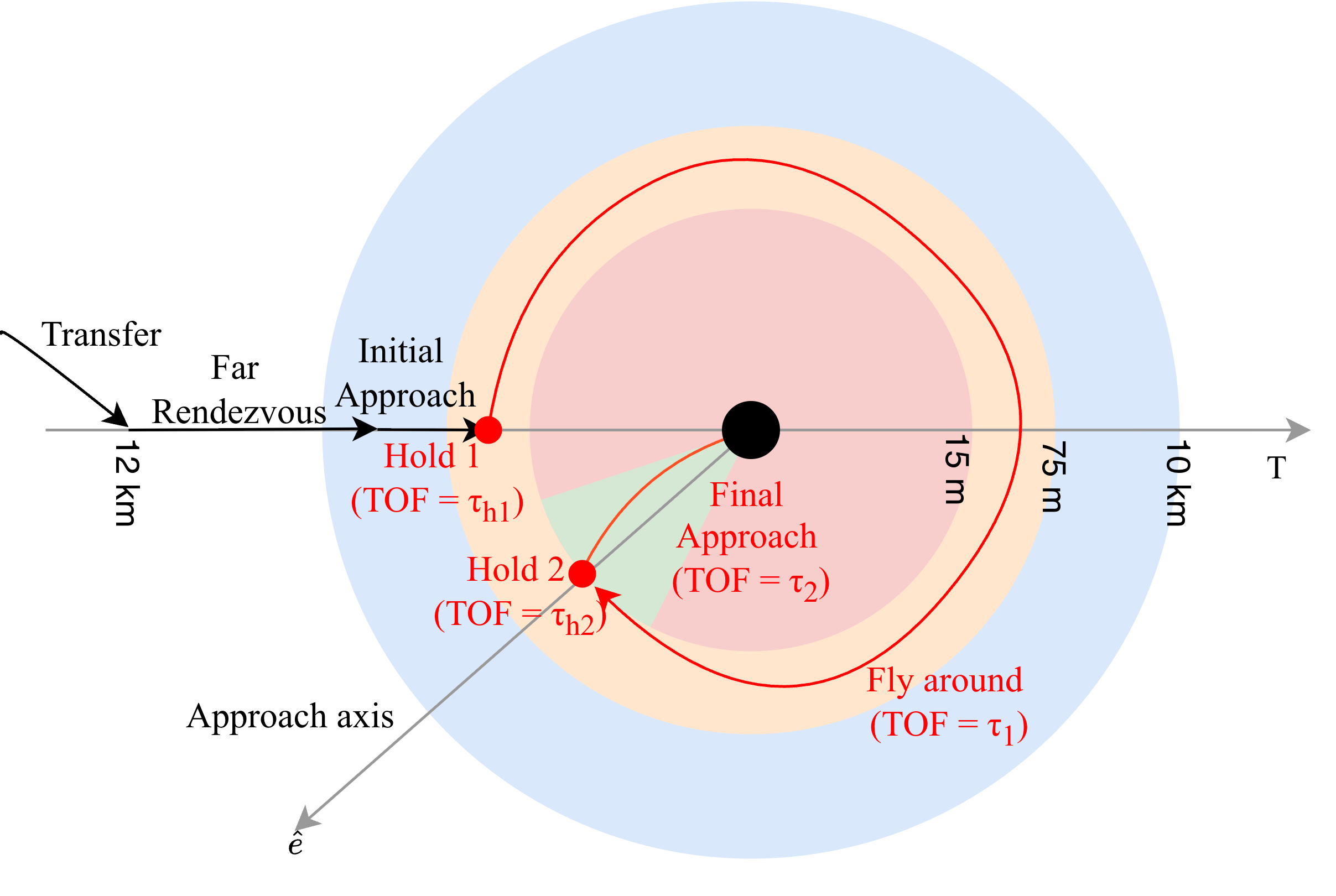}
    \caption{Mission concept of operations and spatial boundaries approached (Close-range rendezvous shown in red. Blue: \ac{RS} (RS), orange: \ac{AS} (AS), red: \ac{KOS}.)}
    \label{fig:VolumesAndZones}
     
\end{figure}

While the full \ac{ISAM} mission consists of launch, transfer, far rendezvous, initial approach, and close-range rendezvous as shown in Fig. \ref{fig:VolumesAndZones}, this work focuses on the close-range rendezvous phase. The close-range rendezvous consists of the fly around phase, approach docking port phase, and additional abort and hold phases; defined as follows
\begin{itemize}[itemsep=0.5ex]
  \item \textbf{Fly around:} The Servicer Spacecraft performs manoeuvres around the \ac{CSO} to reach the docking axis.
    \item \textbf{Approach docking port:} The Servicer Spacecraft performs manoeuvres to make a close-range rendezvous to the \ac{CSO}. The Servicer Spacecraft only enters the \ac{KOS} once it is within the predefined Approach Corridor.
    \item \textbf{Abort:} Abort can be triggered at any proximity phase to autonomously disengage and move the Servicer to a safe location.
    \item \textbf{Hold:} Hold allows the Servicer to maintain a fixed relative state for re-evaluation or replanning.
\end{itemize}

\subsection{Problem Dynamics}
The Servicer and the \ac{CSO} must be propagated over time to simulate true spacecraft motion and attitude changes. This section is divided into translational dynamics and rotational dynamics. Note that the orbital parameters of the \ac{CSO} are denoted by $\Box_{CSO}$ and that of the Servicer are denoted  $\Box_{Serv}$.

\subsubsection{Translational Dynamics}\label{sssec:transdyn}
\paragraph{Relative motion}
For close-proximity operations around a target in a circular reference orbit, the \ac{CW} equations provide a linearized model of relative motion:
\begin{equation}\label{CWdyn}
\ddot{\boldsymbol{r}}^{rel} =
\begin{bmatrix}
3n_{CSO}^2x^{rel} + 2n_{CSO}\dot{y}^{rel} \\
-2n_{CSO}\dot{x}^{rel} \\
-n_{CSO}^2z^{rel}
\end{bmatrix}
+ \frac{1}{m_{Serv}}\boldsymbol{f}^{rel},
\end{equation}

\noindent where \( \boldsymbol{r}^{rel} = [x^{rel}, y^{rel}, z^{rel}]^T \) is the relative position vector, { \( n_{CSO} = \sqrt{\mu_{\oplus}/	\mathfrak{a}_{CSO}^3} \) is the mean motion of the \ac{CSO}, $\mathfrak{a}_{CSO}$ is the semi-major axis of the \ac{CSO}, } \( m_{Serv} \) is the mass of the Servicer, and \( \boldsymbol{f}^{rel} \) represents external control or interaction forces. The parameter $\mu_{\oplus} = \SI{3.986e14}{\meter\cubed\per\second\squared}$ is the standard gravitational parameter of Earth.

To support real-time guidance and optimization, a discretized state-space representation of the \ac{CW} dynamics, obtained by integrating Eq.\eqref{CWdyn} over a fixed time step \( \Delta t \) is used in this work. The discrete-time evolution of the relative state at time step $k$ \( \boldsymbol{x}^{rel}_k = [\boldsymbol{r}^{rel}_k, \boldsymbol{v}^{rel}_k]^T \) is expressed as:
\begin{equation}\label{CWdiscrete}
\boldsymbol{x}^{rel}_{k+1}
= \boldsymbol{\Phi}_{cw}(\Delta t)\left(\boldsymbol{x}^{rel}_k +
\begin{bmatrix}
\boldsymbol{0}_{3\times 1}\\
\Delta\boldsymbol{v}^{rel}_k
\end{bmatrix}\right).
\end{equation}

where \( \Delta \boldsymbol{v}^{rel}_k \) is the impulsive control applied at time step \( k \), and \( \boldsymbol{\Phi}_{cw}(\Delta t)\in \mathbb{R}^{6 \times 6} \) is defined as
\begin{equation}
   \boldsymbol{\Phi}_{cw}(\Delta t) = \begin{bmatrix}
4-3c & 0 & 0 & \frac{s}{n_{CSO}} & \frac{2(1-c)}{n_{CSO}} & 0\\
6(s-n\Delta t) & 1 & 0 & \frac{2(c-1)}{n_{CSO}} & \frac{4s-3n\Delta t}{n_{CSO}} & 0\\
0 & 0 & c & 0 & 0 & \frac{s}{n_{CSO}}\\
3n_{CSO}s & 0 & 0 & c & 2s & 0\\
6n_{CSO}(c-1) & 0 & 0 & -2s & 4c-3 & 0\\
0 & 0 & -n_{CSO}s & 0 & 0 & c
\end{bmatrix}
\end{equation}
where $c = \cos{(n_{CSO}\Delta t)}$ and $s = \sin{(n_{CSO}\Delta t)}$. In this work, Eq. \eqref{CWdiscrete} is used in the convex optimization-based trajectory planning (Section \ref{ssec:ReferenceTrajectoryGeneration}) and convex guidance (Section \ref{stguidance}).

\paragraph{Absolute Motion} For simulating realistic spacecraft motion, a high-fidelity absolute motion model that considers orbital drag and perturbations is utilized in this work. This model is used to simulate true spacecraft motion in the forward propagation step of \ac{CORTEX} in Section \ref{algconvexfprop1}. The equation of motion for both the Servicer and \ac{CSO} is given by

\begin{equation}\label{dynamicsAbs}
\ddot{\boldsymbol{r}} = \boldsymbol{a}_{\text{grav}} + \boldsymbol{a}_{\text{drag}} + \boldsymbol{a}_{\text{3rd-body}} + \frac{1}{m}\boldsymbol{f},
\end{equation}
where  $\boldsymbol{r} = [x, y, z]^T$ is the position vector in an inertial frame. The gravitational acceleration term $\boldsymbol{a}_{\text{grav}}$ is calculated using the Grace gravity model GGM03S \cite{ries2016development} truncated to the 100th degree and order. The acceleration due to the atmospheric drag is given by
\begin{equation}
\boldsymbol{a}_{\text{drag}} = -\frac{1}{2} \frac{C_D A \rho  v^2}{m}\hat{\boldsymbol{v}},
\end{equation}
where $C_D$ is the drag coefficient of the spacecraft, $A$ is the cross-sectional area, $\rho$ is the atmospheric density calculated using NRLMSISE-00 \cite{picone2002nrlmsise}, and $\boldsymbol{v}$ is the spacecraft velocity and {$v = 
\parallel {\boldsymbol{v}} \parallel$.} $\boldsymbol{a}_{\text{3rd-body}}$ represents the third body perturbations from the moon and the Sun.

\paragraph{Conversion from absolute to relative state}
At the end of each forward-propagation step, the inertial (absolute) states of the Servicer and \ac{CSO} are converted to the target-centric \ac{RTN} relative state $\boldsymbol{x}^{rel}$ as follows to compute tracking error with respect to the nominal trajectory.
\begin{equation}\label{eq:abs2rel_block}
\boldsymbol{x}^{rel} =
\begin{bmatrix}
\boldsymbol{C}_{ECI\rightarrow RTN} & \boldsymbol{0}_{3\times 3}\\
\boldsymbol{\Omega}\,\boldsymbol{C}_{ECI\rightarrow RTN} & \boldsymbol{C}_{ECI\rightarrow RTN}
\end{bmatrix}
\left(\boldsymbol{x}_{Serv}-\boldsymbol{x}_{CSO}\right),
\end{equation}
where 
\begin{equation}\label{eq:C_ECI_RTN}
\boldsymbol{C}_{ECI\rightarrow RTN} =\begin{bmatrix} \frac{\boldsymbol{r}_{CSO}}{\parallel \boldsymbol{r}_{CSO} \parallel } \\ 
\frac{\boldsymbol{r}_{CSO} \times \boldsymbol{v}_{CSO} }{\parallel \boldsymbol{r}_{CSO} \times \boldsymbol{v}_{CSO} \parallel } \times \frac{\boldsymbol{r}_{CSO}}{\parallel \boldsymbol{r}_{CSO} \parallel }\\ 
\frac{\boldsymbol{r}_{CSO} \times  \boldsymbol{v}_{CSO} }{\parallel \boldsymbol{r}_{CSO} \times \boldsymbol{v}_{CSO} \parallel }
\end{bmatrix}
\end{equation}
and 
\begin{equation}\label{eq:Omega_def}
\boldsymbol{\Omega}
=
\frac{\|\boldsymbol{r}_{\ac{CSO}}\times\boldsymbol{v}_{\ac{CSO}}\|}{\|\boldsymbol{r}_{\ac{CSO}}\|^{2}}
\begin{bmatrix}
0 & 1 & 0\\
-1 & 0 & 0\\
0 & 0 & 0
\end{bmatrix}.
\end{equation}

\subsubsection{Rotational Dynamics}
Each spacecraft is modelled as a rigid body. The rotational equations of motion are
\begin{equation}\label{eq:rot_dyn}
\mathbf{J}\dot{\boldsymbol{\omega}} + \boldsymbol{\omega}\times(\mathbf{J}\boldsymbol{\omega}) = \boldsymbol{\tau}_{act} + \boldsymbol{\tau}_{env},
\end{equation}
where \(\mathbf{J}\) is the body-frame inertia matrix, \(\boldsymbol{\omega}\) is body angular velocity, \(\boldsymbol{\tau}_{act}\) are actuator torques (e.g., reaction wheels and/or thruster torques), and \(\boldsymbol{\tau}_{env}\) are environmental disturbance torques.
In the simulations presented in this work, \(\boldsymbol{\tau}_{env}\) is neglected unless otherwise stated; attitude dynamics are driven primarily by actuator torques and the thruster moment arms defined in the spacecraft model.

For kinematics, a nonsingular attitude parameterisation is used in simulation and control. When expressed via unit quaternions \(\mathbf{q}=[q_0,q_1,q_2,q_3]^\top\),
\begin{equation}\label{eq:quat_kin}
\dot{\mathbf{q}} = \frac{1}{2}\mathbf{Q}(\boldsymbol{\omega})\mathbf{q},
\end{equation}
with
\begin{equation}
\mathbf{Q}(\boldsymbol{\omega})=
\begin{bmatrix}
0 & -\omega_x & -\omega_y & -\omega_z\\
\omega_x & 0 & \omega_z & -\omega_y\\
\omega_y & -\omega_z & 0 & \omega_x\\
\omega_z & \omega_y & -\omega_x & 0
\end{bmatrix}.
\end{equation}
Attitude enters the \ac{CORTEX} execution pipeline to satisfy pointing objectives such as keeping the target within the camera field-of-view during proximity operations and to realise translational thrust commands through body-fixed actuation.

\section{CORTEX Methodology}\label{ssec:TrajDesg}

\begin{figure*}[h]
    \centering
    \includegraphics[width = 0.9\textwidth]{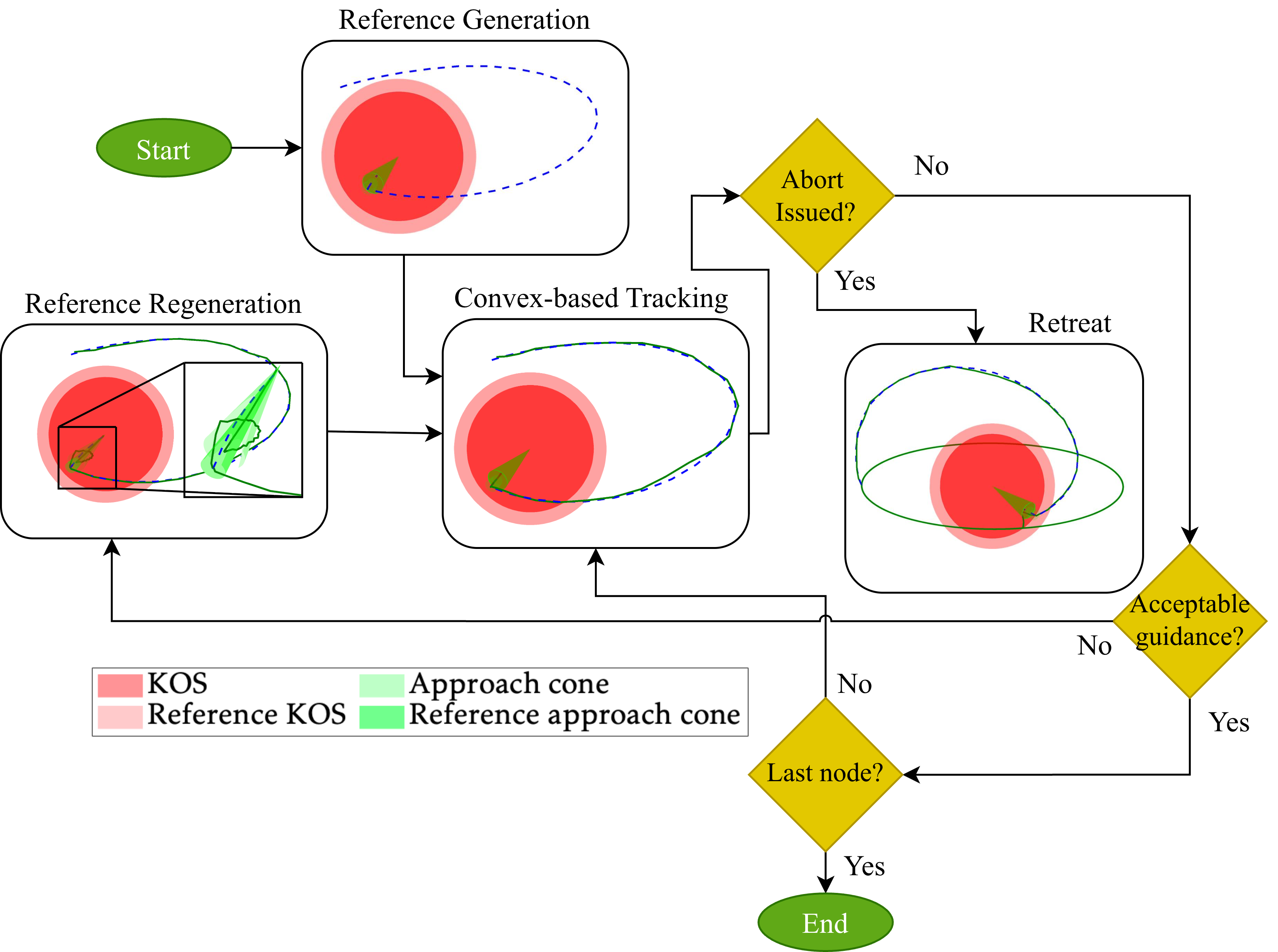}
    \caption{High-level \ac{CORTEX} guidance framework}
    \label{fig:architecturehigh}
\end{figure*}

The primary goals of the trajectory design and guidance in the close-range phase are considered to be: (i) minimize fuel and time consumption, (ii) ensure that both the fly-around and docking occur entirely in sunlight, (iii) avoid the \ac{KOS} during the fly-around phase, (iv) satisfy the approach corridor constraint and avoid plume impingement during the final approach.
Figure~\ref{fig:architecturehigh} summarizes the proposed CORTEX methodology to achieve these goals. It comprises the following four steps.

\begin{enumerate}[itemsep=0.5ex]
    \item \textbf{Reference Generation (Section \ref{ssec:ReferenceTrajectoryGeneration})}, where a time and fuel optimal reference trajectory is generated.
    \item  \textbf{Convex-based Adaptive Tracking (Section \ref{ssec:AdaptiveTracking})}, {where the reference is segmented into equal time sections, and single iteration convex optimization is used to calculate the thrust required to track the reference.  Then, the Servicer is forward propagated under the optimised thrust and realistic, nonlinear dynamics with uncertainties to emulate real spacecraft motion.}
     
    \item \textbf{Reference Regeneration (Section \ref{refregen})}, where if the Servicer deviates significantly from the original reference and violates safety constraints, a new reference is recomputed. 
    
    \item \textbf{Abort and Retreat (Section \ref{abort}) }, where if an abort command is issued by mission control, a retreat manoeuvre is executed to enter a passively safe orbit around the \ac{CSO}.
\end{enumerate}

\subsection {Reference Generation}\label{ssec:ReferenceTrajectoryGeneration}
Firstly, a reference trajectory is generated from the end of the initial approach to the docking port using a hierarchical optimization framework that combines nonlinear and convex optimization techniques.  The outer layer of the reference generation problem consists of a \ac{NP} that determines the optimal \ac{TOF}s for the fly-around phase (\( \tau_{1} \)) and the final approach phase (\( \tau_{2} \)). To ensure that these manoeuvres occur entirely in sunlight, two strategic hold times are introduced: \( \tau_{h_1} \), inserted before the fly-around phase, and \( \tau_{h_2} \), inserted between the fly-around and final approach phases.

The \ac{NP} entails minimizing 
\begin{mini!}
  {\tau_1,\, \tau_2} 
  {J = \sum_{i=1}^{2} (\tau_i + \tau_{hi} + \epsilon_{\mathcal{C}} \mathcal{C}_i)}
  {\label{nonlinearob}} 
  {} 
  \addConstraint{\tau_{lb} \leq \tau_1 \leq \tau_{ub}, \quad \tau_{lb} \leq \tau_2 \leq \tau_{ub}}{}
\end{mini!}

\noindent where $\tau_{lb}$ and $\tau_{ub}$ are the lower and upper bounds of times $\tau_1$ and $\tau_2$. Within each iteration of the \ac{NP}, sequential convex optimization is used to compute fuel-optimal trajectories for fly-around in $\tau_1$ time and final approach in $\tau_2$ time. The fly-around phase problem is denoted as $CP_1$, and the final approach problem as $CP_2$. $\mathcal{C}_1$ and $\mathcal{C}_2$ are constants related to the convergence of $CP_1$ and $CP_2$; if the convex problems successfully converge, they are set to zero, otherwise 1. $\epsilon_\mathcal{C}$ is a weighting factor imposed on that convergence.

\begin{figure}[hbt!]
    \centering
    \includegraphics[width=0.7\linewidth]{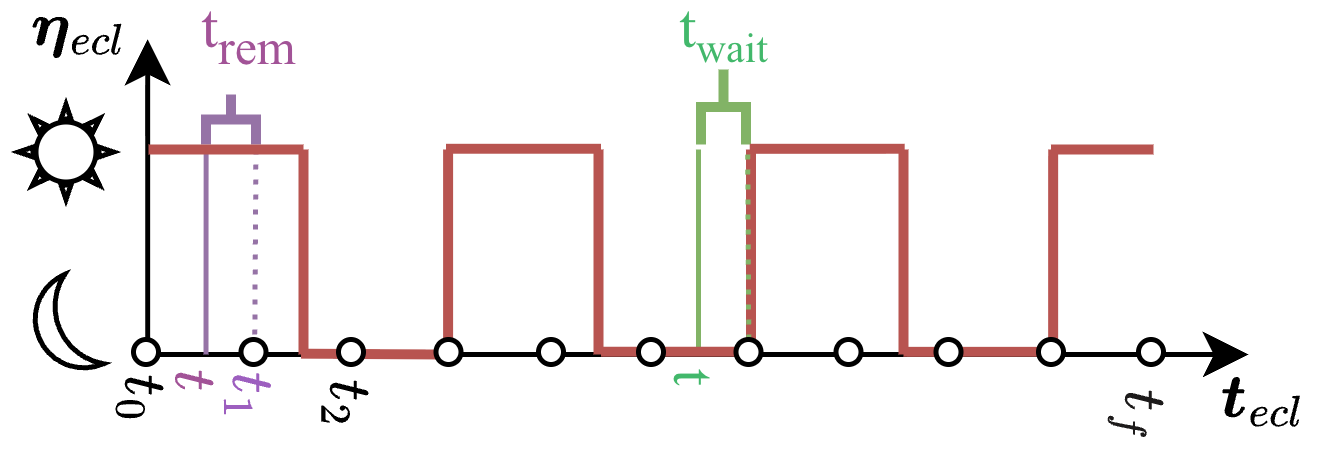}
    \caption{$t_{\text{rem}}$ and $t_{\text{wait}}$ calculation (Note that the eclipse time grid is coarsely sampled in this figure; in practice, a grid with 1000 points per orbit is utilized)}
    \label{fig:eclipses}
\end{figure}

Prior to executing the \ac{NP} routine, the eclipse profile of the \ac{CSO}  \( \boldsymbol{\eta}_{ecl} \) is computed over a discrete time window \( \boldsymbol{t}_{ecl} = \{t_0, t_1, \dots, t_f\} \), where \( t_0 \) is the mission start time and $t_f = t_0 + \tau_1 + \tau_2 + \tau_{{h_1}} 
+ \tau_{{h_2}}$, using the eclipse calculation method described in~\cite{doi:10.2514/1.G008089}. Then, during each objective function evaluation, this eclipse profile is interpolated to determine the current eclipse state at time $t$, denoted $\hat{\eta}_{\text{ecl}}(t)$, where $\hat{\eta}_{\text{ecl}}(t) = 1$ indicates that the \ac{CSO} is in sunlight. If $\hat{\eta}_{\text{ecl}}(t) = 0$ at the start of the fly-around or approach phase, the Sevicer must wait till the next sunlight window. Also if the remaining sunlight window is less than the time required for the transfer, the Servicer must wait till the next Sunlight window begins. The calculation of these wait times are detailed in Figure \ref{fig:eclipses}.

Firstly, the wait time required if \( \hat{\eta}_{\text{ecl}}(t) = 0 \) at the start of the fly-around or final approach phase is computed as the time until the first transition from eclipse to sunlight:
\begin{equation}\label{twait1}
    t_{\text{wait}} = t_{\text{ecl},k} - t
\quad \text{where} \quad
k = \min \left\{ j > i \mid \boldsymbol{\eta}_{\text{ecl},j} = 1 \right\}
\end{equation}
Here, \( i \) is the index such that \( t_{\text{ecl},i} \leq t < t_{\text{ecl},i+1} \). Thus $k$ is the first index of $\boldsymbol{t}_{\text{ecl}}$ after $t$ where $\boldsymbol{\eta}_{ecl,k} = 1$.

Secondly,  \( \hat{\eta}_{\text{ecl}}(t) = 1 \), the remaining time of sunlight \( t_{\text{rem}} \) is computed by linearly scanning forward in the eclipse profile \( \boldsymbol{\eta}_{\text{ecl}} \) from the current time until the first transition to eclipse, and summing the durations of the contiguous sunlight interval such that
\begin{equation}\label{trem}
    t_{\text{rem}} =\left( t_{\text{ecl},i+1} - t \right) +  \sum_{j = i}^{k - 1} (t_{\text{ecl},j+1} - t_{\text{ecl},j}) \cdot \mathbb{I}(\eta_{\text{ecl},j} = 1),
\end{equation}
 \text{where} 
$k = \min \left\{ j > i \mid \eta_{\text{ecl},j} = 0 \right\}$. 
and \( \mathbb{I}(\cdot) \) is the indicator function. If $\tau \geq t_{rem}$, the Servicer must wait till the next sunlight period. In this case: 
\begin{equation}\label{twait2}
t_{\text{wait}} =
\begin{cases}
0, 
& \text{if } t_{\text{rem}} \ge \tau, \\[6pt]
t_{\text{rem}} + \left( t_{\text{ecl},m} - t_{\text{ecl},k} \right),
&  \text{otherwise.}
\end{cases}
\end{equation}

where 
$m = \min \left\{ j \ge k \mid \eta_{\text{ecl},j} = 1 \right\},$ and $k = \min \left\{ j > i \mid \eta_{\text{ecl},j} = 0 \right\}.$ Note that the \(t_{ecl}\) grid shown in the Figure \ref{fig:eclipses} is coarser than the 1000-point-per-orbit discretisation used in this calculation. As such, using Eq.~\eqref{twait1} and Eq.\ref{twait2} to estimate the required wait times only underestimates the remaining sunlight time by only \(\pm 5.6\,\mathrm{s}\). The \ac{NP} objective function calculation is provided in detail in Algorithm~\ref{alg1}.

\begin{algorithm}[h]
\caption{\ac{NP} objective function calculation}\label{alg1}
\begin{spacing}{0.9}
\textbf{Input:} $\tau_1$, $\tau_2$, eclipse profile $\boldsymbol{\eta}_{\text{ecl}}, \boldsymbol{t}_{ecl}$, current time $t$
\begin{algorithmic}[1]
\State $\hat{\eta}_{\text{ecl}}(t) \gets \text{interp}(\boldsymbol{t}_{\text{ecl}}, \boldsymbol{\eta}_{\text{ecl}}, t)$
\If{$\hat{\eta}_{\text{ecl}}(t) = 1$}
    \State Compute $t_{\text{wait}}$ using Eq.\eqref{twait2}; set $\tau_{h_1} \gets t_{\text{wait}}$.
\Else
       \State Compute $t_{\text{wait}}$ using Eq.\eqref{twait1}; set $\tau_{h_1} \gets t_{\text{wait}}$.
\EndIf
\State Solve $CP_1$ with $\tau_1$; set $C_1 \gets \mathbb{I}(CP_1 \text{ fails})$ 
\State $\hat{\eta}_{\text{ecl}} \gets \text{interp}(\boldsymbol{t}_{\text{ecl}}, \boldsymbol{\eta}_{\text{ecl}}, t + \tau_1 + \tau_{h_1})$
\If{$\hat{\eta}_{\text{ecl}} = 0$}
       \State Compute $t_{\text{wait}}$ using Eq.\eqref{twait2}; set $\tau_{h_2} \gets t_{\text{wait}}$.
\Else
     \State Compute $t_{\text{wait}}$ using Eq.\eqref{twait1}; set $\tau_{h_2} \gets t_{\text{wait}}$.
\EndIf
\State Solve $CP_2$ with $\tau_2$; set $C_2 \gets \mathbb{I}(CP_2 \text{ fails})$
\State Compute $J$ from Eq.~\eqref{nonlinearob}
\end{algorithmic}
\textbf{Output:} $J$
\end{spacing}
\end{algorithm}

\subsubsection{Fly around reference generation }\label{CP_1}

The fly around problem ($CP_1$) is defined as:  
\begin{mini!}
  {\boldsymbol{a}^{rel}(t)} 
  {\int_{0}^{\tau_1} \|\overline{\boldsymbol{a}}^{rel}(t)\| \, dt} 
  {\label{CP_1obj}}{}
  \addConstraint{\overline{\boldsymbol{x}}^{rel}(0) = \overline{\boldsymbol{x}}^{rel}_{i_1} \quad  \overline{\boldsymbol{x}}^{rel}(\tau_1) = \overline{\boldsymbol{x}}^{rel}_{f_1}}
    \addConstraint{ \text{Eq. \eqref{CWdyn}} \quad  \forall t \in [0, \tau_1]}
  \addConstraint{\|\overline{\boldsymbol{a}}^{rel}(t)\| \leq \overline{a}_{\text{max}}, \quad \forall t \in [0, \tau_1]}
  \addConstraint{\|\overline{\boldsymbol{r}}^{rel}(t)\| \geq \overline{\rho}_{KOS}, \quad \forall t \in [0, \tau_1]}
\end{mini!}
where $\overline{\boldsymbol{x}}_{i_{1}}^{rel}$ and $\overline{\boldsymbol{x}}_{f_{1}}^{rel}$ are the initial and target states of the fly around phase.
Note that $\overline{\vphantom{A}\Box}$ is used to denote states and variables associated with the reference trajectory. 

To preserve conservatism in reference generation and improve robustness during guidance, the available thrust is assumed to be \(20\%\) lower than the true maximum, i.e., \(\overline{a}_{\max}=0.8\,a_{\max}\). Similarly, the \ac{KOS} radius ${\rho}_{KOS}$ is inflated by \(20\%\) relative to its nominal value, \(\overline{\rho}_{\text{KOS}}=1.2\,\rho_{\text{KOS}}\).

The problem can be convexified as follows, optimizing $\Delta \overline{v}^{rel}$ instead of thrust acceleration for convenience.  First, a discretized time vector over $\tau_1$ is denoted as  $\overline{t}_{CP_{1}} = [0: \Delta t_{CP_1} :\tau_1, \tau_1]^T$ where $\Delta t_{CP_1}$ is a user-set timestep. Then, the number of $\Delta v$ nodes is set to be $N_1 =\text{len}(\overline{t}_{CP_{1}})-1$. 
Thus, the convexified $CP_1$ problem becomes: 
\begin{mini!}
  {\{\Delta \overline{\boldsymbol{v}}_k^{rel}\}_{k=1}^{N_1}} 
  {\sum_{k=1}^{N_1} \overline{s}_k} 
  {\label{CP_1obj2}}{}
  \addConstraint{\boldsymbol{x}_1^{rel} = \overline{\boldsymbol{x}}_{i_1}, \quad \overline{\boldsymbol{x}}_{N_1+1}^{rel} = \boldsymbol{x}_{f_1}}
  \addConstraint{\text{Eq.~\eqref{CWdiscrete}}, \quad \forall k \in [1, N_1]}
  \addConstraint{\|\Delta \overline{\boldsymbol{v}}_k^{rel}\|_2 \leq \overline{s}_k, \quad \forall k \in [1,N_1]}
  \addConstraint{ \overline{s}_k \leq \overline{a}_{\max}\,\Delta \overline{t}_{CP_{1}}, \quad  \forall k \in [1,N_1]}
\end{mini!}

Note that the \ac{KOS} constraint is nonconvex, and need to be implemented via statically-attached planes placed at boundary points around the sphere as discussed in \cite{Ortolano2021Autonomous}. These planes, which are convex constraints, are defined as 
\begin{equation}\label{bpcons}
    \overline{\rho}_{KOS} \leq \overline{\boldsymbol{r}}_{k_{v}}^T \overline{\boldsymbol{l}}_{k_{v}}^* \  \quad \forall k_v 
\end{equation}
where $\overline{\boldsymbol{l}}_{k_v}^* =  \overline{\boldsymbol{r}}_{k_v}^* /\parallel  \overline{\boldsymbol{r}}_{k_v}^*\parallel$ and ${k_v}$ is the array of nodes that violate the \ac{KOS} constraint when the convex problem is solved without considering Eq. \eqref{bpcons}. The process for solving $CP_1$ with the \ac{KOS} constraint is the adapted from \cite{Ortolano2021Autonomous}, and entails:
\begin{enumerate}
    \item Solve \( CP_1 \) without \ac{KOS} constraint in Eq.~\eqref{bpcons}.
    \item Setup the boundary plane constraints as shown in Eq.~\eqref{bpcons} at points that violate the \ac{KOS} constraint in the previous solution.
    \item Solve \( CP_1 \) with Eq.~\eqref{bpcons}.
    \item Repeat steps 2 and 3 till convergence.
\end{enumerate}
Using these steps,  $CP_1$ can be solved via successive convex optimization. The obtained reference trajectory and $\Delta v$ control solution are denoted as
$\overline{\boldsymbol{x}}_{CP_{1}}$ and  $\Delta \overline{\boldsymbol{v}}_{CP_{1}}$, respectively.

\subsubsection{Final approach reference generation}\label{CP_2}
\begin{figure}[hbt]
    \centering
    \includegraphics[width=0.6\linewidth]{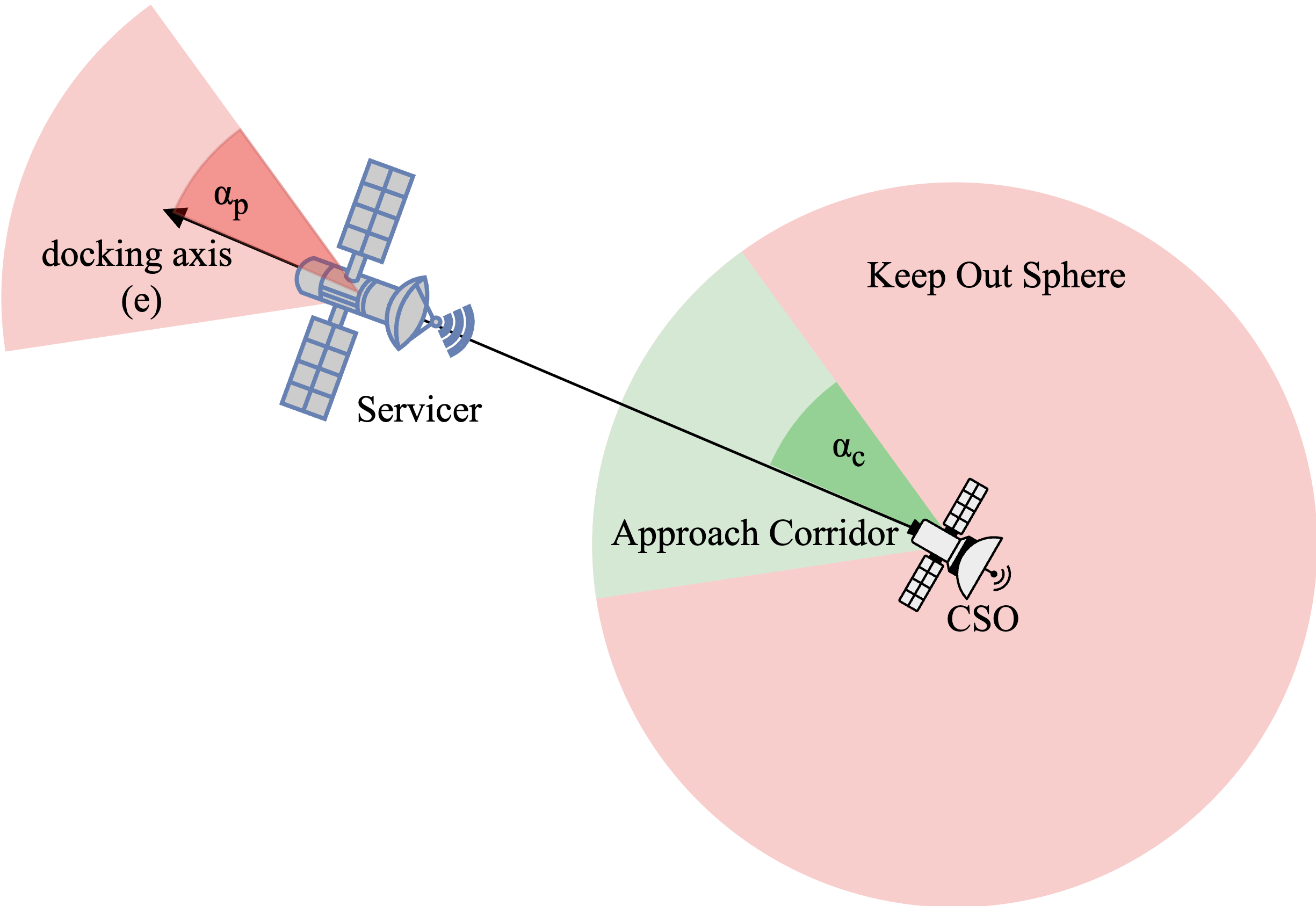}
    \caption{Docking dynamics}
    \label{fig:dockingref}
\end{figure}

The final approach problem ($CP_2$)  aims to optimise the thrust acceleration while adhering to the plume impingement and approach cone constraints. It can be formulated as  
\begin{mini!}
  {\overline{\mathbf{a}}^{rel}(t)} 
  {\int_0^{\tau_2} \|\overline{\mathbf{a}}^{rel}(t)\|\, dt} 
  {\label{CP_2obj}}{}
  \addConstraint{\overline{\mathbf{x}}^{rel}(0) = \overline{\mathbf{\boldsymbol{x}}}_{i_2}, \quad \overline{\mathbf{x}}^{rel}(\tau_2) = \overline{\mathbf{\boldsymbol{x}}}_{f_2}}
  \addConstraint{\|\overline{\mathbf{a}}^{rel}(t)\| \leq \overline{a}_{\text{max}}, \quad \forall t \in [0, \tau_2]}
  \addConstraint{\overline{\mathbf{a}}^{rel}(t) \cdot \overline{\mathbf{r}}^{rel}(t) \leq \|\overline{\mathbf{a}}^{rel}(t)\|\, \|\overline{\mathbf{r}}^{rel}(t)\| \cos (\overline{\alpha}_{p}) {\label{eq:plumeimpingement}}}
     \addConstraint{ \text{Eq. \eqref{CWdyn}} \quad  \forall t \in [0, \tau_2]}
\addConstraint{\|\overline{\mathbf{r}}^{rel}(t)\| \cos(\overline{\alpha}_{c}) \leq \hat{\mathbf{e}}^T \overline{\mathbf{r}}^{rel}(t) {\label{eq:approachcorridor}}}
\end{mini!}

where for all \( t \in [0, \tau_2] \), the states \( \overline{\mathbf{\boldsymbol{x}}}_{i_2} =\overline{\mathbf{\boldsymbol{x}}}_{f_1} \) and \( \overline{\mathbf{\boldsymbol{x}}}_{f_2} \) denote the initial-and the end point of the fly around phase- and target conditions for the final approach phase, respectively. Again, to ensure a conservative and robust guidance framework, reduced thrust authority, a more restrictive plume impingement constraint, and a narrower approach corridor are assumed. Specifically, \( \overline{a}_{\text{max}} = 0.8\, a_{\text{max}} \), \( \overline{\alpha}_{p} = 1.2\, \alpha_{p} \), and \( \overline{\alpha}_{c} = 0.5\, \alpha_{c} \), where \( \alpha_{p} \) is the plume impingement angle as defined in~\cite{panag2023plume} and Figure~\ref{fig:dockingref}, and \( \alpha_c \) is the half-angle of the approach corridor. The approach axis is denoted by \( \hat{\boldsymbol{e}} \).

This problem can be convexified as follows, again optimizing $\Delta\overline {v}$ instead of thrust acceleration. Once again, a descritized time vector over $\tau_2$ is defined as  $\overline{t}_{CP_{2}} = [0: \Delta t_{CP_2} :\tau_2, \tau_2]^T$ where $\Delta t_{CP_2}$ is a user-set timestep. Then, the number of $\Delta v$ nodes for $CP_2$ is set as $N_2 =\text{len}(\overline{t}_{CP_{2}})-1$. 

Now, the convexified formulation of $CP_2$ is derived as: 
\begin{mini!}
  {\{\Delta \overline{\boldsymbol{v}}_k^{rel}\}_{k=1}^{N_2}} 
  {\sum_{k=1}^{N_2} \overline{s}_k} 
  {\label{CP_2obj2}}{}
  \addConstraint{\boldsymbol{x}_1^{rel} = \overline{\boldsymbol{x}}_{f_1}, \quad \overline{\boldsymbol{x}}_{N_2+1}^{rel} = \boldsymbol{x}_{f_2}}
  \addConstraint{\text{Eq. \eqref{CWdiscrete}}, \quad \forall k \in [1, N_2]}
  \addConstraint{\|\Delta \overline{\boldsymbol{v}}_k^{rel}\| \leq \overline{s}_k, \quad \forall k}
  \addConstraint{   \overline{s}_k \leq \overline{a}_{\text{max}} \cdot \Delta \overline{t}_{CP_{2}}, \quad \forall k}
  \addConstraint{\|\overline{\boldsymbol{r}}_k^{rel}\| \cos(\overline{\alpha}_{c}) \leq \hat{\boldsymbol{e}}^T \overline{\boldsymbol{r}}_k^{rel}, \quad \forall k}
\end{mini!}
\vfill

The constraint given in Eq. \eqref{eq:plumeimpingement} is the plume impingement constraint, which restricts the direction of thrust acceleration such that the angle between $\overline{\boldsymbol{a}}^{rel}(t)$ and $\overline{\boldsymbol{r}}^{rel}(t) $ always exceeds $\overline{\alpha}_{p}$. This constraint is discussed in detail in \cite{panag2023plume}. In this work, we simplify this constraint by assuming fixed servicer orientation. Handling this constraint in conjunction with attitude control is left as future work. As Eq. \eqref{eq:plumeimpingement} is non-convex, successive convexification is utilized to solve $CP_2$ with it, adapting the strategy discussed in \cite{panag2023plume}. Here, Eq. \eqref{eq:plumeimpingement} is first discretized as 

\begin{equation}
    h_k = \Delta \overline{\boldsymbol{v}_k}^{rel} \cdot \overline{\mathbf{r}}^{rel}_k - \|\overline{\boldsymbol{v}_k}^{rel} \|\, \|\overline{\mathbf{r}}^{rel}_k\| \cos (\overline{\alpha}_{p})  \leq 0
\end{equation}
Then it is linearized to become

\begin{equation}\label{limpimp}
h_{k}(\overline{\boldsymbol{y}}^m) + \nabla h_{k}(\overline{\boldsymbol{y}}^m)(\overline{\boldsymbol{y}}^{m+1} - \overline{\boldsymbol{y}}^m) \leq 0
\end{equation}
for all $N \in [0, N_2]$, where, $\overline{\boldsymbol{y}}^m = [\overline{\boldsymbol{x}}^{rel,m}, \Delta \overline{\boldsymbol{v}}^{rel,m}, \overline{s}^k]$ is the $m$th iteration of the successive convexification. Then, the overall steps for solving $CP_2$ and obtaining the reference trajectory for final approach are:
\begin{enumerate}[itemsep=0.5ex]
    \item Solve $CP_2$ without Eq. \eqref{limpimp}.
    \item Introduce the linearized plume impingement constraint given in Eq. \eqref{limpimp} and solve $CP_2$, using solution from step 1 as a guess.  
    \item Repeat step 2 till convergence.
\end{enumerate}

The reference trajectory and the $\Delta v$ control obtained is denoted $\overline{\boldsymbol{x}}_{CP_{2}}^{rel}$ and  $\Delta \overline{\boldsymbol{v}}_{CP_{2}}^{rel}$, respectively. 

\subsubsection{Summary}
The \ac{NP} problem given in Eq. \eqref{nonlinearob} can be solved using a nonlinear optimization method such as \ac{PSO}\cite{kennedy1995particle},  where $CP_1$ and $CP_2$ are solved repeatedly under different $\tau_1$ and $\tau_2$ values. This results in a time optimized robust reference trajectory 
$\overline{\boldsymbol{x}}^{rel} = [\overline{\boldsymbol{x}}^{rel}_{CP_1},\overline{\boldsymbol{x}}^{rel}_{CP_2}]^T$ and a reference $\Delta v$ control  profile $\Delta \overline{\boldsymbol{v}}^{rel} = [\Delta \overline{\boldsymbol{v}}^{rel}_{CP_1},\Delta \overline{ \boldsymbol{v}}^{rel}_{CP_2}]^T$ for the time profile $\overline{t} = [\tau_{h_1} , \tau_{h_1}+\overline{t}_{CP_1},\tau_{h_1}+\tau_{1} + \tau_{h_2}, \tau_{h_1}+\tau_{1} + \tau_{h_2}+\overline{t}_{CP_2}]$.

\subsection{Convex-based Adaptive Tracking}\label{ssec:AdaptiveTracking}

Once the reference trajectory has been generated, a convex-based adaptive tracking strategy is employed to guide the Servicer spacecraft along the nominal path under realistic dynamical uncertainties.
Rather than executing the reference open loop, the trajectory is segmented into short tracking intervals, enabling closed-loop guidance and real-time replanning behaviour consistent with on-orbit spacecraft operations.

A high-level summary of the convex-based adaptive tracking strategy is provided in Algorithm~\ref{convextrack}. Within it, $\mathcal{S}$ defines the mission phase, where $h_1$ is the first hold, $1$ is the Fly-around, $h_2$ is the second hold, and $2$ is the Final approach.    Within each mission phase $\mathcal{S}\in\{h_1,1,h_2,2\}$, guidance is executed at a fixed update period $\Delta t_g$.
Let $k$ denote the discrete guidance index, with relative state $\boldsymbol{x}^{rel}_k$ evaluated at the start of guidance step $k$.
The guidance is therefore executed over $N_{g} = \left\lceil \frac{\tau_{1} +\tau_{2} + \tau_{h_1} + \tau_{h_2} }{\Delta t_g} \right\rceil$ guidance steps. A global guidance time grid is defined as
\begin{equation}
\boldsymbol{t}^{guid}
=
\{ t^{guid}_0,\ t^{guid}_1,\ \ldots,\ t^{guid}_{N_g} \},
\qquad
t^{guid}_k = t^{guid}_0 + k\,\Delta t_g,\quad k = 0,\ldots,N_g ,
\end{equation}
where the final guidance interval may satisfy $t^{guid}_{N_g}-t^{guid}_{N_g-1} \le \Delta t_g$
when the total execution duration is not an integer multiple of $\Delta t_g$.

At each step, a convex guidance problem $CP_g$ is solved to compute an impulsive control increment that steers the current state $\boldsymbol{x}^{rel}_k$ toward a terminal reference state $\overline{\boldsymbol{x}}^{rel}_{T,k}$ over a single guidance interval of duration $\Delta t_g$.
During transfer phases ($\mathcal{S}\in\{1,2\}$), $\overline{\boldsymbol{x}}^{rel}_{T,k}$ is obtained by interpolation of the nominal reference trajectory; during hold phases ($\mathcal{S}\in\{h_1,h_2\}$), it is set to the station-keeping target.

Following the computation of the guidance control, the Servicer spacecraft and the \ac{CSO} are propagated forward using the high-fidelity translational dynamics presented in Section~\ref{sssec:transdyn}.
This propagation includes realistic modelling effects, including state estimation errors and thrust execution errors, thereby emulating true spacecraft motion between guidance updates.

The adaptive nature of the tracking scheme arises from continuous monitoring of constraint satisfaction and tracking performance.
If the deviation between the true relative state and the reference exceeds a prescribed threshold, or if safety constraints such as keep-out-zone violation, approach corridor infringement, or plume impingement are detected, the tracking process is terminated and a new reference recomputation is triggered or the mission is aborted. 

A mission abort command is issued for the three conditions defined in Table \ref{tab:abort_recompute}. If none of the abort conditions are satisfied, the recompute conditions in Table \ref{tab:abort_recompute} are checked. These are less stringent than the abort criteria and are based on the violation of the buffered \ac{KOS}, plume impingement and approach corridor conditions.

\begin{table*}[hbt!]
\centering
\small
\caption{Abort and Recompute Trigger Conditions}
\label{tab:abort_recompute}
\renewcommand{\arraystretch}{1.35}
\begin{tabular}{p{1.1cm} p{6.6cm} p{6.8cm}}
\hline
 & {Description} & {Logical Definition}  \\
\hline
\multicolumn{3}{c}{{Mission Abort Conditions}} \\
\hline
$\mathcal{C}_{\text{\ac{KOS}}}$
&
Violation of the true keep-out zone during hold or fly-around phases.
&
$\mathcal{S} \in \{h_1,h_2,1\}
\ \text{and}\
\|\boldsymbol{r}^{rel}_{k+1}\| < r_{\mathrm{\ac{KOS}}}$
\\

$\mathcal{C}_{\text{plume}}$
&
Plume impingement on the target during final approach.
&
$\mathcal{S}=2
\ \text{and}\
\boldsymbol{a}^{rel}_{k+1}\!\cdot\!\boldsymbol{r}^{rel}_{k+1}
>
\|\boldsymbol{a}^{rel}_{k+1}\|\,\|\boldsymbol{r}^{rel}_{k+1}\|
\cos(\alpha_p)$
\\

$\mathcal{C}_{\text{cone}}$
&
Violation of the approach corridor geometry during final approach.
&
$\mathcal{S}=2
\ \text{and}\
\hat{\boldsymbol{e}}^\top \boldsymbol{r}^{rel}_{k+1}
<
\|\boldsymbol{r}^{rel}_{k+1}\|\cos(\alpha_c)$
\\

\hline
\multicolumn{3}{c}{{Mission Recompute Conditions}} \\
\hline

$\mathcal{C}_{\text{trk}}$
&
Excessive deviation from the reference trajectory.
&
$\|\boldsymbol{r}^{rel}_{k+1}
- \overline{\boldsymbol{r}}^{rel}_{k+1}\|
>
\epsilon_{r_{\mathcal{S}}}$
\\

$\overline{\mathcal{C}}_{\text{\ac{KOS}}}$
&
Violation of buffered keep-out-zone used for guidance feasibility.
&
$\mathcal{S} \in \{h_1,h_2,1\}
\ \text{and}\
\|\boldsymbol{r}^{rel}_{k+1}\| < \overline{r}_{\mathrm{\ac{KOS}}}$
\\

$\overline{\mathcal{C}}_{\text{cone}}$
&
Violation of the buffered approach cone constraint.
&
$\mathcal{S}=2
\ \text{and}\
\hat{\boldsymbol{e}}^\top \boldsymbol{r}^{rel}_{k+1}
<
\|\boldsymbol{r}^{rel}_{k+1}\|
\cos(\overline{\alpha}_c)$
\\

$\overline{\mathcal{C}}_{\text{plume}}$
&
Violation of buffered plume impingement constraint.
&
$\mathcal{S}=2
\ \text{and}\
\boldsymbol{a}^{rel}_{k+1}\!\cdot\!\boldsymbol{r}^{rel}_{k+1}
>
\|\boldsymbol{a}^{rel}_{k+1}\|\,\|\boldsymbol{r}^{rel}_{k+1}\|
\cos(\overline{\alpha}_p)$
\\

\hline
\end{tabular}
\end{table*}

If none of these violations are encountered, the convex-tracking is repeated segment-by-segment till the end of the final approach. 

\begin{algorithm*}[hbt!]
\caption{Convex-based Adaptive Tracking}
\label{convextrack}
\begin{spacing}{0.9}
\textbf{Input:} phase durations $\tau_1$, $\tau_2$, $\tau_{h_1}$, $\tau_{h_2}$;
reference time grid $\overline{t}$;
reference states $\overline{\boldsymbol{x}}^{rel}$;
initial relative state $\boldsymbol{x}^{rel}(t_0)$;
guidance step $\Delta t_g$;
convex substep $\delta t_g$;
recomputation thresholds $\epsilon_{r_\mathcal{S}}$.
\begin{algorithmic}[1]
\State Initialise global time $t \gets 0$
\For{each phase $\mathcal{S} \in \{h_1, 1, h_2, 2\}$}
    \If{$\tau_{\mathcal{S}} > 0$}
        \State $N_{g} \gets \left\lceil (\tau_1 + \tau_2 + \tau_{h_1} + \tau_{h_2})/\Delta t_g \right\rceil$
        \If{$\mathcal{S} \in \{h_1, h_2\}$}
            \State Set station-keeping target $\overline{\boldsymbol{x}}^{rel}_{T} \gets \overline{\boldsymbol{x}}^{rel}_{CP_{\mathcal{S}}}$
        \EndIf
        \For{$k = 1$ to $N_{g}$}
            \If{$\mathcal{S} \in \{1, 2\}$}
                \State $\overline{\boldsymbol{x}}^{rel}_{T} \gets
                \text{interp}(\overline{t}, \overline{\boldsymbol{x}}^{rel}, t^{guid}_{k+1})$
            \EndIf
            \State Solve $CP_g$ with $N_{\text{sub}}$ substeps to obtain $\{\Delta \boldsymbol{v}^{rel}_j\}_{j=1}^{N_{\text{sub}}}$ (Section~\ref{stguidance})
            \State Propagate $\boldsymbol{x}^{rel}_k$ using Alg.~\ref{algconvexfprop} to obtain
            $\boldsymbol{x}_{k+1}^{rel} = [\boldsymbol{r}_{k+1}^{rel}, \boldsymbol{v}_{k+1}^{rel}]^\top$
            \If{$\mathcal{C}_{\text{\ac{KOS}}} \ \textbf{or} \ \mathcal{C}_{\text{plume}} \ \textbf{or} \ \mathcal{C}_{\text{cone}}$}
                \State \textbf{return} \texttt{Abort}
            \EndIf
            \If{$\mathcal{C}_{\text{trk}} \ \textbf{or} \ \overline{\mathcal{C}}_{\text{\ac{KOS}}} \ \textbf{or} \ \overline{\mathcal{C}}_{\text{plume}} \ \textbf{or} \ \overline{\mathcal{C}}_{\text{cone}}$}
                \State \textbf{return} \texttt{Recompute}
            \EndIf
        \EndFor
    \EndIf
\EndFor
\end{algorithmic}
\end{spacing}
\end{algorithm*}

\subsubsection{Convex guidance problem} \label{stguidance}


In this work, the continuous-time guidance problem $CP_{g}$ is formulated as
\begin{mini!}
  {\boldsymbol{a}^{rel}(t)}
  {\int_{\boldsymbol{t}^{guid}_k}^{\boldsymbol{t}^{guid}_{k+1}} \|\boldsymbol{a}^{rel}(t)\|\, dt
  + \left\| \Delta \boldsymbol{x}^{rel}_f \right\|}
  {\label{cg1}}{}
  \addConstraint{\boldsymbol{x}^{rel}(t_k) = \boldsymbol{x}^{rel}_k}
  \addConstraint{\text{Eq.~\eqref{CWdyn}}, \quad \forall t \in [\boldsymbol{t}^{guid}_{k},\, \boldsymbol{t}^{guid}_{k+1}]}
  \addConstraint{\|\boldsymbol{a}^{rel}(t)\| \leq a_{\text{max}}, \quad \forall t}
  \addConstraint{\|\boldsymbol{r}^{rel}(t)\|\cos(\alpha_c)
  \leq \hat{\boldsymbol{e}}^\top \boldsymbol{r}^{rel}(t),
  \quad \forall t \ \text{if } \mathcal{S}=2}
\end{mini!}
where the terminal deviation is defined as
$\Delta \boldsymbol{x}^{rel}_f
= \boldsymbol{x}^{rel}(t_{i+1}) - \boldsymbol{x}^{rel}_T. $

To obtain a tractable optimisation problem, the formulation in Eq.~\eqref{cg1} is convexified. At each guidance update time $t^{\text{guid}}_k$, the time interval
$[t^{\text{guid}}_k,\ t^{\text{guid}}_{k+1}]$ is discretised for the convex
optimisation into $N_{\text{sub}}$ uniform segments. The corresponding local time grid is defined as 

\begin{equation}
\boldsymbol{t}^{\text{cvx}}_k
=
\left\{
t^{\text{cvx}}_{k,1},\ t^{\text{cvx}}_{k,2},\ \ldots,\ t^{\text{cvx}}_{k,N_{\text{sub}}+1}
\right\},
\end{equation}
with
\begin{equation}
t^{\text{cvx}}_{k,j}
=
t^{\text{guid}}_k
+
(j-1)\frac{t^{\text{guid}}_{k+1}-t^{\text{guid}}_k}{N_{\text{sub}}},
\qquad
j = 1,\ldots,N_{\text{sub}}+1.
\end{equation}

This results in the following second-order cone program:
\begin{mini!}
  {\{\Delta \boldsymbol{v}^{rel}_j\}_{j=1}^{N_{\text{sub}}}}
  {\sum_{j=1}^{N_{\text{sub}}} s^{rel}_j
  + \epsilon_{\mathcal{P}}\epsilon_{x_f}}
  {\label{eq:subtracking}}{}
  \addConstraint{\boldsymbol{x}^{rel}_1 =\boldsymbol{x}^{rel}_k}
  \addConstraint{\text{Eq.~\eqref{CWdiscrete}},
  \quad \forall j = 1,\ldots,N_{\text{sub}}}
  \addConstraint{\|\Delta \boldsymbol{v}^{rel}_j\| \le s^{rel}_j,
  \quad \forall j}
  \addConstraint{\|\Delta \boldsymbol{v}^{rel}_j\|
  \le a_{\max}\,\Delta t_g,
  \quad \forall j}
  \addConstraint{\left\|
  \boldsymbol{x}^{rel}_{N_{\text{sub}}+1}
  - \overline{\boldsymbol{x}}^{rel}_f
  \right\|
  \le \epsilon_{x_f}}
  \addConstraint{\|\boldsymbol{r}^{rel}_j\|\cos(\alpha_c)
  \le \hat{\boldsymbol{e}}^\top \boldsymbol{r}^{rel}_j,
  \quad \forall j = 1,\ldots,N_{\text{sub}}+1}
\end{mini!}

The approach cone constraint (final line) is enforced only when $\mathcal{S}=2$.
The scalar $\epsilon_{\mathcal{P}}$ denotes a user-defined penalty weight
associated with soft terminal constraint satisfaction. Solving the convex
programme yields the sequence of impulsive velocity increments
$\Delta \boldsymbol{v}^{rel}_j$, which guides the spacecraft from the current
relative state $\boldsymbol{x}^{rel}_k$ toward the desired terminal state
$\overline{\boldsymbol{x}}^{rel}_T$, enforced as a soft constraint.

Non-convex constraints, including plume impingement and \ac{KOS} are not explicitly included in the convex guidance formulation to
preserve single-iteration convergence. Instead, these constraints are handled during reference trajectory generation through conservative margins on the plume
half-angle $\alpha_p$ and the \ac{KOS} radius $r_{\text{\ac{KOS}}}$. If violations are
detected, the reference trajectory is recomputed. The explicit integration of
non-convex constraints into the convex guidance problem is left for future work.

\subsubsection{Forward propagation for a convex tracking segment }\label{algconvexfprop1}

Finally, after obtaining $\Delta\boldsymbol {v}^{rel}$ from the previous section,the two spacecraft are propagated under high-fidelity dynamics,  state and thrust errors as well as missed thrust events. This process is given in Algorithm \ref{algconvexfprop}. First, missed thrust events are introduced by randomly cancelling all scheduled impulses with probability
$p_{\text{misthrust}}$. In addition, thrust execution errors are modelled through perturbations in impulse magnitude and direction, while state uncertainty is captured via additive position and velocity injection errors at each convex time step. At every step, the perturbed relative state is converted to Cartesian coordinates, propagated under high fidelity dynamics in the absolute frame, and subsequently
mapped back to the relative frame. This forward propagation provides a
realistic assessment of tracking performance and generates the initial
condition for the subsequent guidance update.

\begin{algorithm}[hbt!]
\caption{Forward Propagation for Convex Tracking Segment}
\label{algconvexfprop}
\begin{spacing}{0.9}
\textbf{Input:} $\boldsymbol{x}^{rel}_k$, $\{\Delta \boldsymbol{v}^{rel}_j\}_{j=1}^{N_{\text{sub}}}$ , $\boldsymbol{t}^{cvx}_{k}$, $\boldsymbol{x}_{CSO}$.
\begin{algorithmic}[1]
\State Introduce missed thrust events by turning thrust off randomly.  Draw $M \sim \mathcal{N}(0,1)$ and let 
\begin{equation}
    \Delta \boldsymbol{v}^{rel} \gets 
\begin{cases}
\boldsymbol{0}, & \text{if } M < p_{\text{misthrust}} \\
\Delta \boldsymbol{v}^{rel}, & \text{otherwise}
\end{cases}
\end{equation}

\For{$j = 1$ to $\text{len}(\boldsymbol{t}^{cvx}_k)$}
    \State Inject State Errors: 
    $\boldsymbol{r}^{rel} \gets \boldsymbol{r}^{rel} + \Delta r \cdot  \frac{\boldsymbol{\delta}_{r^{rel}}}{\sqrt{3}},
        \boldsymbol{v}^{rel} \gets \boldsymbol{v}^{rel} + \Delta v \cdot  \frac{\delta_{v^{rel}}}{\sqrt{3}}$
    where $\boldsymbol{\delta}_{r^{rel}} \sim \mathcal{N}(0, \sigma_{r^{rel}})$, $\delta_{v^{rel}} \sim \mathcal{N}(0, \sigma_{v^{rel}})$.
    \State {Inject thrust errors to the thrust magnitude and angles.}
    \begin{equation}
         \Delta v_{j_E} = \Delta {v}_j (1 + \delta_{\Delta v}), \  \beta_{j_E} = \beta_j + \delta_\beta \  \text{and}   \ \alpha_{j_E} = \alpha_j + \delta_\alpha
    \end{equation}
    where $    \beta_j = \sin^{-1}( \frac{v_{j,n}}{\Delta {v}_j })$ and $\alpha_j = \tan^{-1}( \frac{v_{j,t}}{v_{j,n}})$. The errors are drawn from $\delta_{\Delta v} \sim \mathcal{N}(0, \sigma_{\Delta v})$, 
     $\delta_\beta \sim \mathcal{N}(0, \sigma_\beta)$, and $\delta_\alpha \sim \mathcal{N}(0, \sigma_\alpha)$.
    
    \State Compute Perturbed $\Delta v$ as 
              $\Delta \boldsymbol{v}_{j_E} = \Delta v_{j_E}[c_{\beta_{j_E}}s_{\alpha_{j_E}}, c_{\beta_{j_E}} c_{\alpha_{j_E}},  s_{\beta_{j_E}} ]^T$,
    where $s_{\theta}/c_{\theta} = \sin{\theta} / \cos{\theta}$.
    
    \State {Update $\boldsymbol{x}^{rel}$ and convert to absolute coordinate:}
    \begin{align}
          \boldsymbol{x}^{rel} &\gets \boldsymbol{x}^{rel} + [\boldsymbol{0}_{3\times1};\ \Delta \boldsymbol{v}_{j_E}]\\
           \boldsymbol{x}_{Serv} &\gets \text{Rel2Cart}(\boldsymbol{x}^{rel}, \boldsymbol{x}_{CSO}) 
    \end{align}

    \State Propagate $\boldsymbol{x}_{Serv}$, $\boldsymbol{x}_{CSO}$ over $\Delta t(j \rightarrow j+1)$ via Eq.~\eqref{dynamicsAbs}.
    \State Update $\boldsymbol{x}^{rel}_{k+1} \gets \text{Cart2Rel}(\boldsymbol{x}_{Serv}, \boldsymbol{x}_{CSO})$
\EndFor
\end{algorithmic}
\end{spacing}
\end{algorithm}


\subsection{Reference Regeneration}\label{refregen}

\begin{figure}[tbp]
    \centering
    \includegraphics[width=\linewidth]{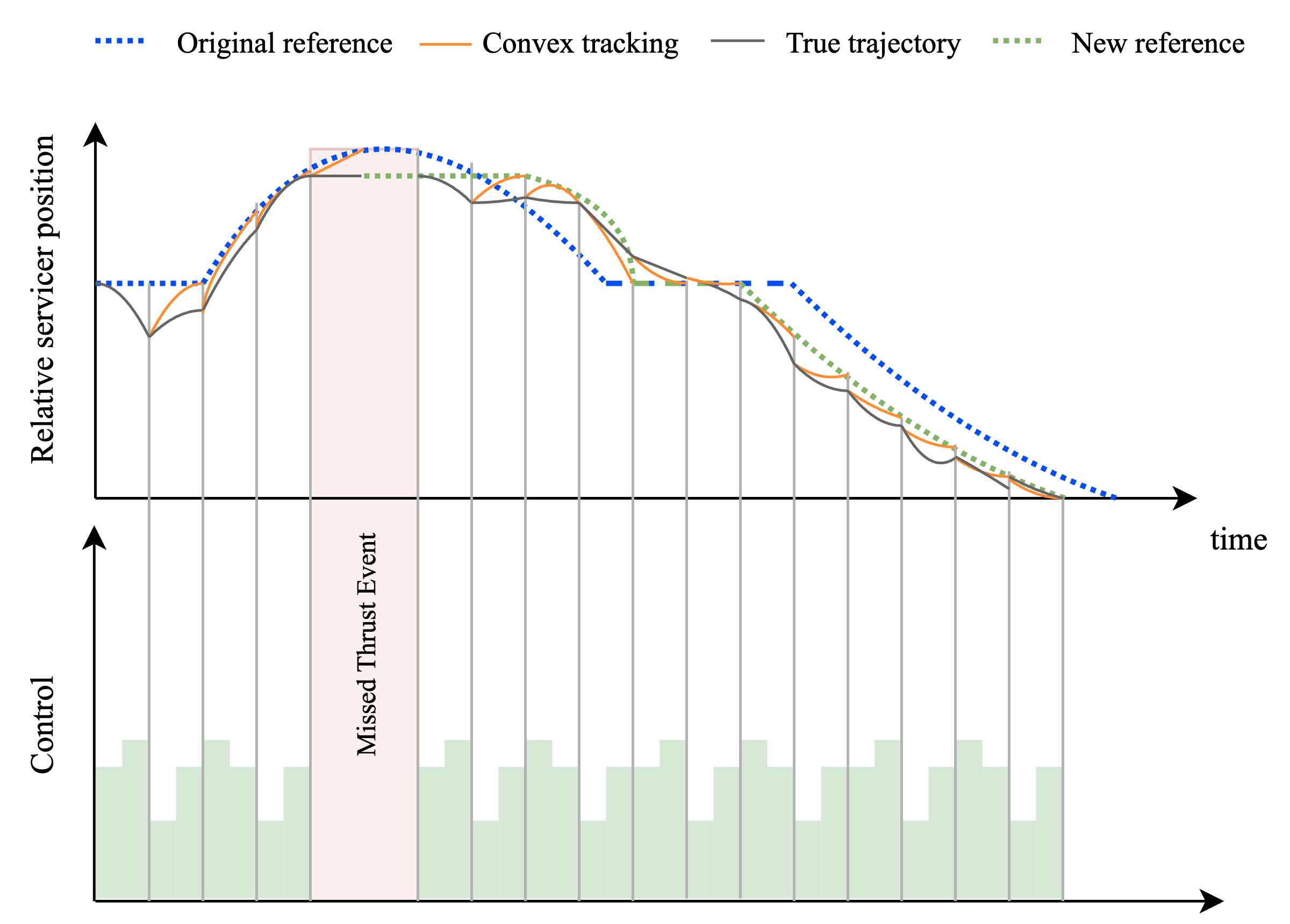}
    \caption{Recomputation due a missed thrust event during the fly around phase (No abort command issued).}
    \label{fig:mtRecomp}
\end{figure}

When a recomputation is triggered, the relative velocity is immediately nullified, and the process discussed in Section \ref{ssec:ReferenceTrajectoryGeneration} is repeated to generate a new reference from the deviated point \( \boldsymbol{x}_{k+1}^{\text{rel}}\) to the docking port  as shown in Figure \ref{fig:mtRecomp}. Then, tracking is conducted as discussed in section \ref{ssec:AdaptiveTracking}. 




\subsection{Abort and Retreat}\label{abort}
\begin{figure}[tbp]
    \centering
    \includegraphics[width=1\linewidth]{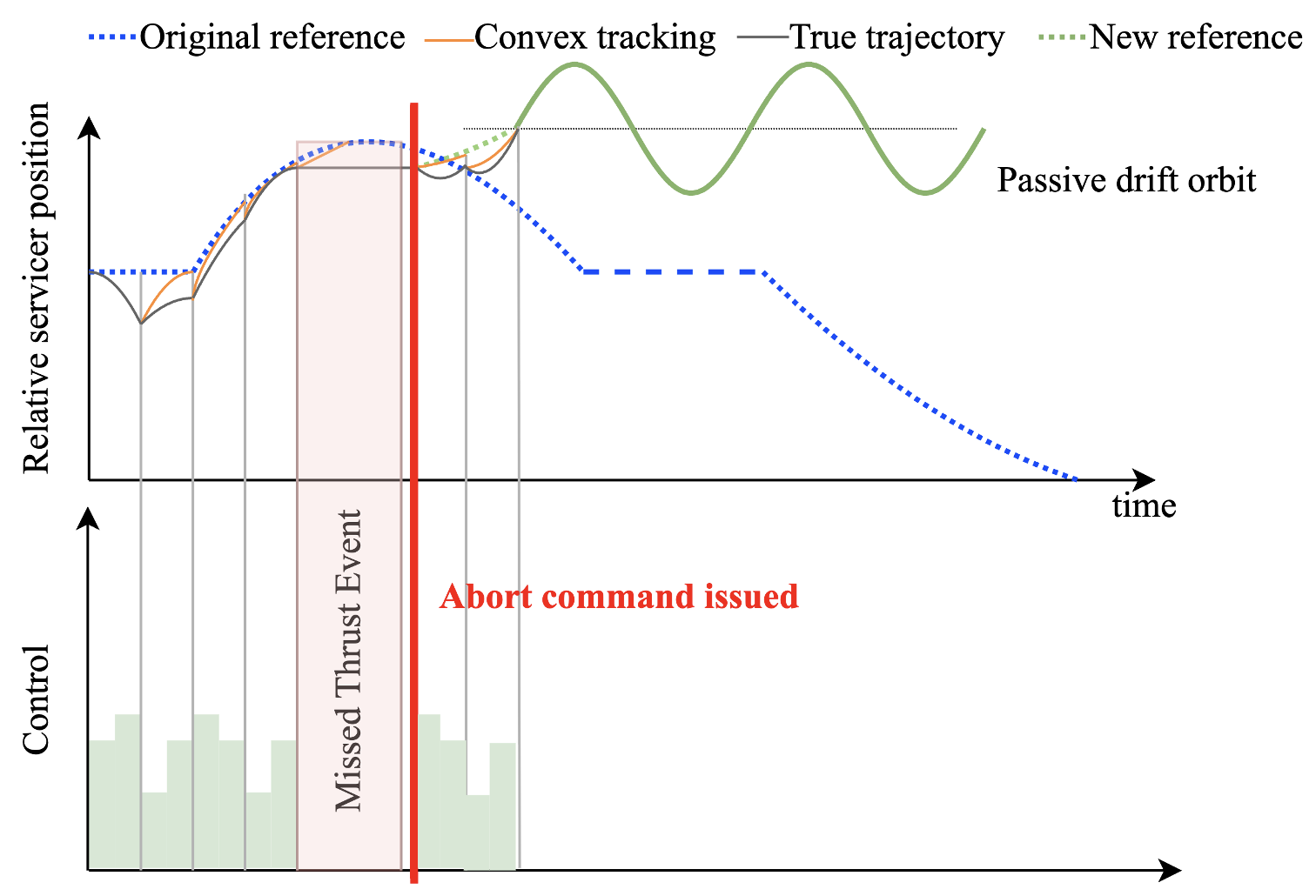}
    \caption{Abort and retreat after a missed thrust event}
    \label{fig:abr}
\end{figure}

At any stage during the close-range rendezvous, if an abort command is issued from the ground, the spacecraft nullifies its current relative velocity and retreats to a circumnavigating elliptical orbit around the \ac{CSO}, located outside the \ac{KOS}. The same reference generation and guidance strategy described earlier is used to reach a designated safe orbit, defined by the conditions \( \dot{x} = \frac{n_{\text{CSO}}}{2} y \), \( \dot{y} = 0 \), and \( y = \frac{r_{\text{AS}}}{2} \). This procedure is illustrated in Figure~\ref{fig:abr}. This is the last step of the CORTEX architecture.


\section{Experimental Testbed Setup}

\subsection{Software Simulation Testbed}

\begin{figure*}[ht]
    \centering
    \includegraphics[width=\textwidth]{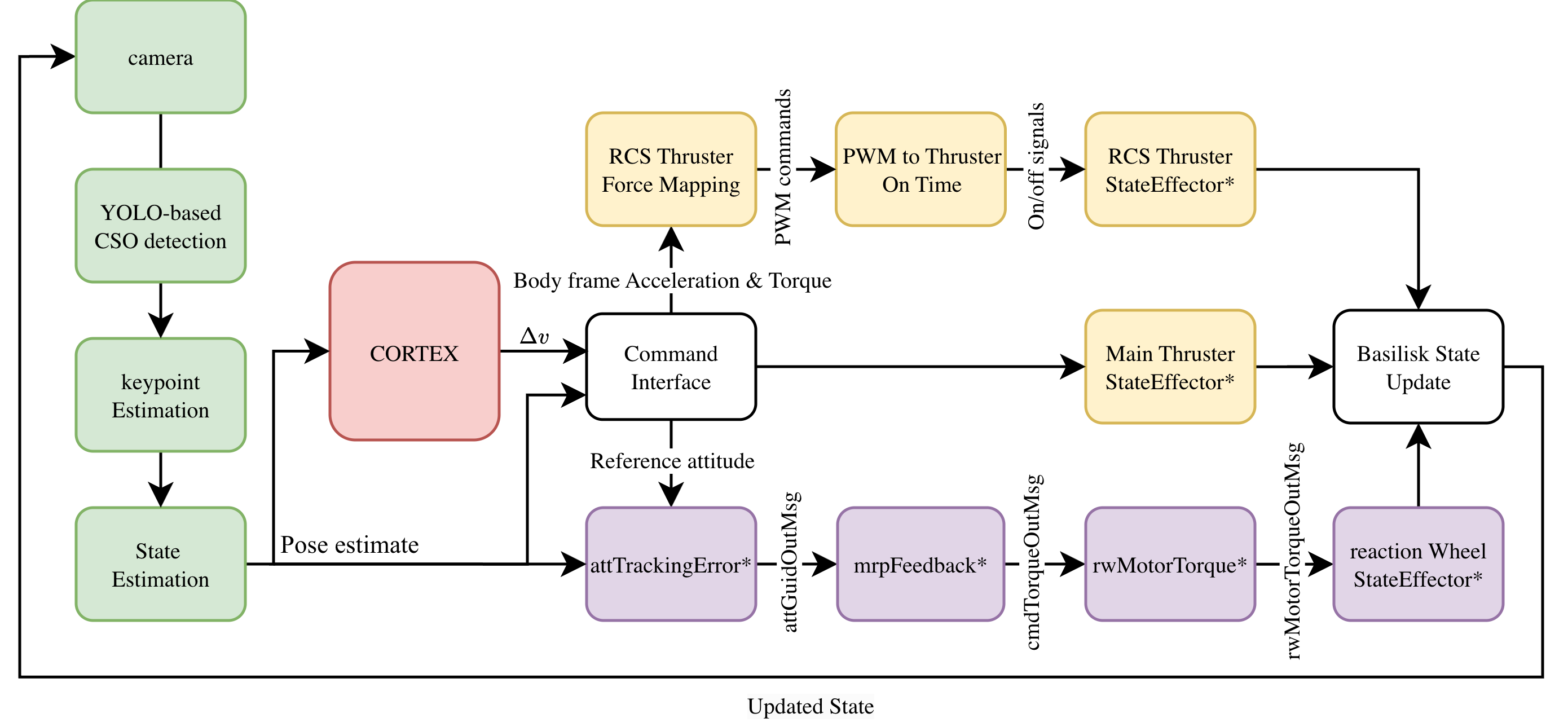}
    \caption{Overview of the Basilisk software testbed. Colour coding indicates functional layers: perception (green), CORTEX autonomy (red), attitude determination and control (purple), and thruster control (yellow). Asterisks (*) denote standard Basilisk modules.}
    \label{fig:SstUmlControl}
\end{figure*}

To evaluate the proposed guidance algorithm under realistic conditions in a software setting, the Basilisk astrodynamics simulation framework was employed \cite{doi:10.2514/1.I010762}. Basilisk enables modular simulation of spacecraft subsystems, and is used here to model both the Servicer and the \ac{CSO} with sufficient fidelity to validate control performance under realistic conditions.
This section focuses on the control architecture implemented on the Servicer spacecraft. Note that a high-level overview of the broader simulation setup including navigation, mission logic, and additional subsystems can be found in \cite{guinane2025iac}. Figure~\ref{fig:SstUmlControl} illustrates the relevant Basilisk modules used in the control loop, as well as their interconnections. 

The major modules present in the software testbed are:
\begin{enumerate}[itemsep=0.5ex]
      \item \textbf{Perception:} Processes onboard imagery to estimate the relative pose of the \ac{CSO} during close-range operations.
    
    \item \textbf{CORTEX:} CORTEX generates initial trajectory plans and $\Delta v$ guidance commands based on the current and desired spacecraft states.
    \item \textbf{Attitude Determination and Control:} These modules command the reaction wheels to achieve the desired attitude.
    \item \textbf{Thruster Control:} These modules command both the main thruster and the \ac{RCS} thrusters to execute orbital manoeuvres and perform translational control.
\end{enumerate}

\subsubsection{Perception Module}

During close-range rendezvous, relative state information of the \ac{CSO} is obtained
through a vision-based perception pipeline that operates in three sequential stages.
\begin{itemize}
    \item \textbf{\ac{YOLO} for CSO detection}: A compact convolutional object detector based on the \ac{YOLO} architecture \cite{Jocher2023} is used to
localize the \ac{CSO} in each monocular grayscale image taken.
\item \textbf{Key Point Estimation}:The
detected region of interest is passed to a keypoint-based feature
estimation network, which predicts the 2-D coordinates of a predefined set of
structural keypoints on the \ac{CSO}.
\item \textbf{State Estimation}: Batch initialization procedure is used to obtain an initial estimate of the relative
position, velocity, attitude, and rotation rate of the \ac{CSO}. Then visual measurements are recursively fused over time using a dynamic estimation model to reduced state uncertainty. 
\end{itemize}
Note that the detailed network architectures, training procedures, and quantitative perception
performance will be provided in the companion perception study~\cite{Illa2026PerceptionDataset}.


\subsubsection{CORTEX Module}
This module implements the CORTEX methodology and consists of a trajectory planner and guidance module that work together to generate, track, and execute reference trajectories.



\subsubsection{Command Interface Module}
This module implements the command interface between high-level $\Delta v   $ outputs from CORTEX and the low-level Basilisk attitude and thruster control stack. It reads the $\Delta\mathbf{v}$ commands, and produces an attitude reference to orient the Servicer towards the CSO and propulsion commands for the main engine and/or the \ac{RCS}. For main-engine manoeuvres, the module computes a reference attitude that aligns the body-fixed thrust axis with the required inertial manoeuvre direction, and converts the planned $\Delta v$s into a main-thruster on-time $t_{\mathrm{on}}$ using the rocket equation such that 
\begin{equation}
t_{\mathrm{on}}
=
\frac{m_0 - m_f}{\dot{m}},
\end{equation}
where 
$m_f = m_0\exp\!\left(-\frac{\|\Delta\mathbf{v}\|}{I_{sp}g_0}\right)$
and 
$\dot{m}=\frac{F}{I_{sp}g_0}$. $F$ is the thrust magnitude, $I_{sp}$ is the specific impulse, and $m_0$ is the current spacecraft mass.
For RCS execution, the planned $\Delta\mathbf{v}$ in the active guidance window is converted into an inertial acceleration command $\mathbf{a}_N$ that achieves the desired velocity change within the window duration. 



\subsubsection{Thruster Control Modules}

The thruster control system is composed of four main modules. 
\begin{itemize}
    \item \textbf{RCS Thruster Force Mapping Module:}
    Maps desired body frame acceleration and torque commands into \ac{PWM} commands for each \ac{RCS} thruster using a \ac{MILP} formulation as shown in Algorithm \ref{alg:milp_pwm}.

    \item \textbf{PWM to Thruster On Time Module}:
    Converts continuous \ac{PWM} values into discrete On/Off signals for each \ac{RCS} thruster, where the on-time for each thruster is computed as $t_{on,i} = dt_i \cdot t_\text{control}$.  Then the On-off signal is generated as 
    \begin{equation}
            \text{OnTime}_i(t) = \begin{cases}
               1.0 & \text{if } (t \bmod t_{control}) < t_{on,i} \\
               0.0 & \text{otherwise}
            \end{cases}
        \end{equation}
        
    \item \textbf{RCS Thruster and Main Thruster StateEffector Modules}: These are standard basilisk modules \footnote{\url{https://hanspeterschaub.info/basilisk/Documentation/simulation/dynamics/Thrusters/thrusterStateEffector/thrusterStateEffector.html}} that execute the thruster firings based on the computed on-time commands.
\end{itemize}

\begin{algorithm}[tb]
\caption{\ac{MILP}-Based Force and Torque to \ac{PWM} Mapping}
\label{alg:milp_pwm}
\begin{spacing}{0.9}
\begin{algorithmic}[1]
\State \textbf{Input:} Number of thrusters $n$, maximum thrust $F_{\text{max},i}$ for each thruster, desired force/torque vector $\boldsymbol{c} \in \mathbb{R}^6$, control period $t_{\text{control}}$
\State \textbf{Decision Variables:} $dt_i \in [0,1]$ for each thruster $i = 1,...,n$; tracking error $e_j \geq 0$ for $j = 1,...,6$
\State Compute actuation matrix $\boldsymbol{A} \in \mathbb{R}^{6 \times n}$:
\For{$i = 1$ to $n$}
    \State $\boldsymbol{A}_{1:3,i} \gets \hat{\boldsymbol{t}}_{i,B}$ \Comment{Unit thrust vector}
    \State $\boldsymbol{A}_{4:6,i} \gets \boldsymbol{r}_{i,B} \times \hat{\boldsymbol{t}}_{i,B}$ \Comment{Torque contribution}
\EndFor
\State Define \ac{MILP} constraints:
\For{$i = 1$ to $n$}
    \State $0 \leq dt_i \leq 1$ \Comment{\ac{PWM} bounds}
\EndFor
\For{$j = 1$ to $6$}
    \State $\sum_{i=1}^{n} \boldsymbol{A}_{j,i} F_{\text{max},i} dt_i - e_j \leq c_j$
    \State $\sum_{i=1}^{n} \boldsymbol{A}_{j,i} F_{\text{max},i} dt_i + e_j \geq c_j$
\EndFor
\State \textbf{Objective:} $\min \sum_{j=1}^{6} e_j$
\State Solve the \ac{MILP}. 
\State \textbf{Output:} Optimal \ac{PWM} settings $dt_i$ for each thruster
\end{algorithmic}
\end{spacing}
\end{algorithm}

\subsubsection{Reaction Wheel Control Modules}

All reaction wheel control modules utilized already exist within the main Basilisk framework. 
The \texttt{attTrackingError} module computes the attitude tracking error by comparing the current Servicer attitude with the reference attitude, and outputs an attitude guidance message. This message is passed to the \texttt{mrpFeedback} module, which implements a nonlinear attitude tracking controller to compute torque commands in the body frame. These torque commands are then routed through the \texttt{rwMotorTorque} module, which maps them to individual reaction wheel inputs. Finally, the \texttt{reactionWheelStateEffector} module applies the computed torques to propagate the spacecraft attitude dynamics.

\subsubsection{Basilisk State Update}
Lastly, the Basilisk simulation environment propagates the states of both satellites using nonlinear absolute dynamics augmented with environmental perturbations, state and thrust noise. This propagation setup has already been shown in detail in Algorithm \ref{algconvexfprop}. 
The results generated via the software simulation testbed are detailed in Section \ref{softwareRes}. 


 \subsection{Hardware Simulation Testbed }

\begin{figure*}[tbp]
    \centering
    \begin{subfigure}[t]{0.24\linewidth}
        \centering
        \includegraphics{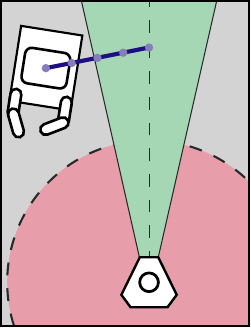}
        \caption{Initial trajectory planned and executed to enter the approach corridor.}
    \end{subfigure}\hfill
    \begin{subfigure}[t]{0.24\linewidth}
        \centering
        \includegraphics{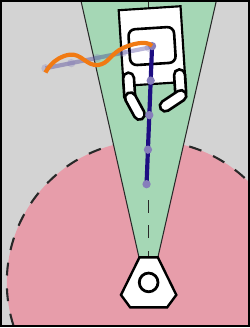}
        \caption{Once on the approach axis, trajectory planned and executed for close-range rendezvous}
    \end{subfigure}\hfill
    \begin{subfigure}[t]{0.24\linewidth}
        \centering
        \includegraphics{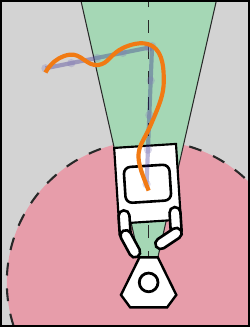}
        \caption{If close-range rendezvous is successful and the pre-capture position is reached, docking begins.}
    \end{subfigure}\hfill
    \begin{subfigure}[t]{0.24\linewidth}
        \centering
        \includegraphics{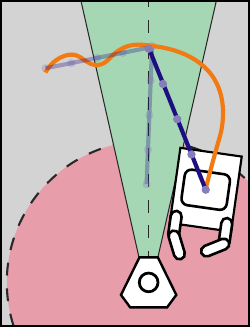}
        \caption{If inside the \ac{KOS}, an abort trajectory is planned and executed to return to the initial close-range rendezvous position.}
    \end{subfigure}
    \caption{Concept of operations for the experimental validation using the hardware testbed. Black/Grey line: Trajectory Plan, Orange line: True trajectory. 
    }
    \label{fig:ExpConOps}
\end{figure*}

\begin{figure*}[tbp]
    \centering    \includegraphics[width = 0.9\textwidth]{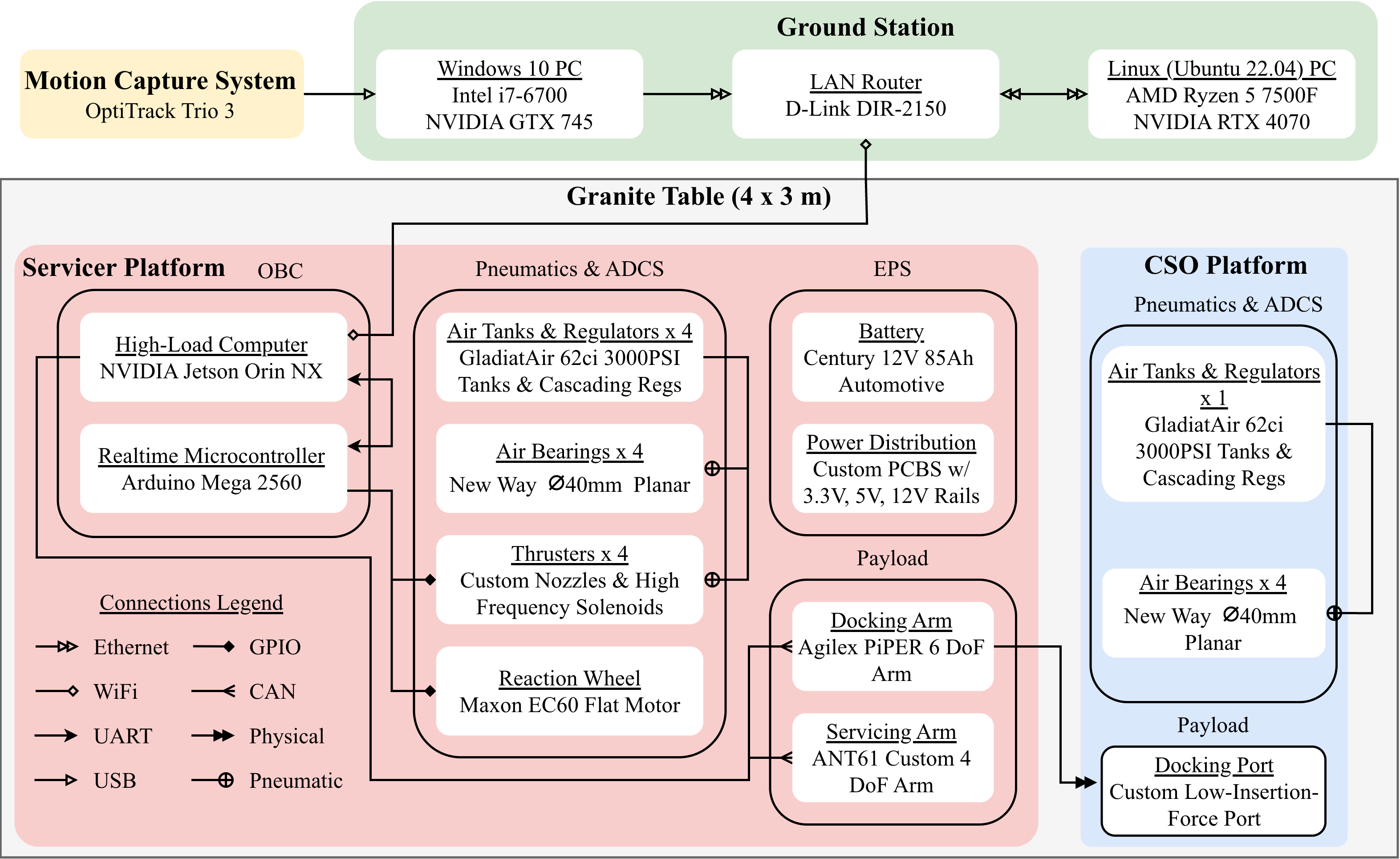}
    \caption{A system diagram of the hardware testbed subsystems. }
    \label{fig:PatSystemDiagram}
\end{figure*}

\begin{figure*}[tbp]
    \centering
    \includegraphics[width=0.8\linewidth]{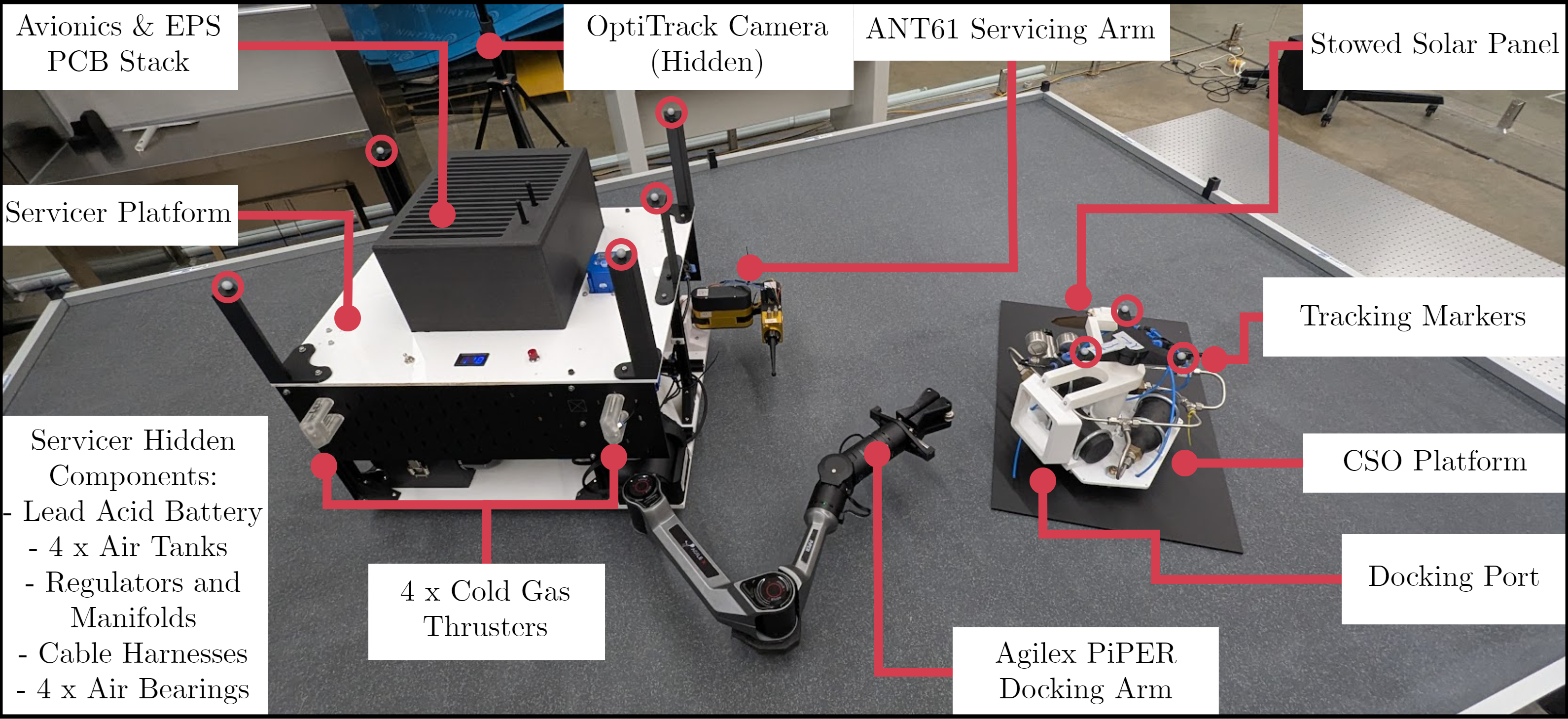}
    \caption{Diagram of the hardware testbed servicer platform and \ac{CSO} platform.}
    \label{fig:PATDiagram}
\end{figure*}

The hardware testbed is used to test a simplified version of the fly-around and close-range rendezvous modes, as shown in Figure \ref{fig:ExpConOps}.
The testbed consists of an active servicer platform and a passive \ac{CSO} platform on a large granite table with a ground station and motion capture, as shown in Figure \ref{fig:PATDiagram}. A system block diagram is also given in Figure \ref{fig:PatSystemDiagram}.

The experimental platforms operate on a $3 \times 2$ m monolithic granite slab with DIN876 grade~000 flatness, corresponding to a surface deviation below $10,\mu$m. The slab is supported by pneumatically isolated vibration-damping legs and is levelled prior to each experiment to reduce uncontrolled drift during free-floating operation. Any residual drift is treated as a bounded external disturbance acting on the platform dynamics and is actively compensated by the onboard controllers.

Platform motion is measured using an OptiTrack Trio~3 \footnote{\url{https://optitrack.com/cameras/trio-3}}  motion capture system, which tracks infrared markers mounted on each vehicle. The system provides position and attitude estimates of each platform’s geometric centre at 120~Hz with sub-millimetre accuracy.

A ground station is used for data aggregation and real-time control computation. A Windows PC runs the OptiTrack Motive software and streams platform pose estimates, while a Linux Ubuntu~22.04 PC records experimental data and executes \ac{ROS2} nodes alongside MATLAB and Simulink-based control modules. 

\subsubsection{Servicer Platform}
The servicer platform has two layers.
The lower layer contains the pneumatics (including tanks, manifolds, valves, and thrusters), battery, and robotic arms.
The upper layer contains the \ac{EPS} and avionics \acp{PCB} (which includes the \ac{OBC}), reaction wheel, switches, and displays.

\paragraph{Pneumatics}
The pneumatics subsystem supplies and controls the air bearings and reaction control thrusters.
The air tanks allow for approximately 30 minutes of floating time on each charge to 3000 PSI.
Each of the thrusters is a custom-designed nozzle with a theoretical maximum thrust of $1.66$ N, controller with solenoid valves with a maximum switching frequency of $500$ Hz. While the platform also has a Maxon EC60 motor for use as a reaction wheel, the experiments presented below only use the reaction control thrusters.

\paragraph{Electrical Power System}
The \ac{EPS} consists of a lead-acid battery that powers all electrical components on the servicer platform at the required voltages.

\paragraph{Robotic Arms}
Two robotic arms are included on the platform. The Agilex PiPER\footnote{\url{https://global.agilex.ai/products/piper}} is used for the docking tasks, along with a custom docking gripper.

\paragraph{OBC \& Software}
The \ac{OBC} is a combination of an NVIDIA Jetson Orin NX\footnote{\url{https://www.nvidia.com/en-us/autonomous-machines/embedded-systems/jetson-orin/}} running Ubuntu 22.04, for high-level, high-computational-load operations and hosting of \ac{ROS2}, and an Arduino Mega 2560 microcontroller running FreeRTOS for low-level sensor recording and actuator control with real-time response capabilities.
The software is defined at three levels: (1) Control layer which performs
integrated control of the servicer platform state, (2) ROS2 Layer for robust interfacing of sensors, actuators, and devices, and (3) Embedded layer with microcontroller code which drives hardware on the servicer platform.

\subsubsection{CSO Platform}
The \ac{CSO} platform is a passive structure that includes a docking port for contactless insertion. The platform also has infrared tracking markers for motion capture.

Together, this experimental infrastructure enables repeatable, closed-loop validation of the CORTEX framework under realistic sensing, actuation, and other constraints. Despite operating in a planar environment, the testbed preserves the key couplings between perception, guidance, and control that arise during spacecraft proximity operations, thereby providing a representative platform for assessing system-level performance prior to on-orbit deployment.

\section{Results}
\subsection{Software Testbed Results}\label{softwareRes}
In this section, software testbed results are provided for the close-range rendezvous phase trajectory generation and guidance under different thrust and state error levels. The orbital parameters of the \ac{CSO} at $t_0$ are $[a,e,i, \Omega, \omega,\theta ]^T = [\SI{6878.1}{\kilo\meter}, 0.001, \SI{98.0}{\degree}, \SI{0.1}{\degree}, \SI{0.1}{\degree},\SI{0.1}{\degree}]^T$. The wet mass of the Servicer is assumed to be $500$ kg, and the minimum total thrust available from the six \ac{RCS} thrusters is 1.2 N, which we consider to be the maximum thrust to be conservative. Hence $a_{max} = 1.2/500 =  \SI{2.4}{\milli\meter\per\second\squared}$. The mission start date $t_0$ is taken to be $2022:05:01$  $00:00:00$ UTC.  Other problem parameters associated with the test cases are given in Table \ref{trajpoints}.


\begin{table*}[tbp]
\centering
\small
\caption{Problem parameters.}
\setlength{\tabcolsep}{5pt}
\begin{tabular}{ll ll}
\toprule
\textbf{Parameter} & \textbf{Value} & \textbf{Parameter} & \textbf{Value} \\
\midrule
$m_{\mathrm{Serv}}$ (kg) & 500
& $a_{\max}^{\mathrm{rel}}$ (mm/s$^2$) & 2.4 \\

$t_0$ (UTC) & $2022{:}05{:}01\ 00{:}00{:}00$
& $[r_{RS},r_{AS},r_{KOS}]$ (m) & [$\SI{10000}{}$, $\SI{75}{}$, $\SI{15}{}$] \\

$[\alpha_p,\alpha_c]$ (deg) & [$\SI{20}{}$, $\SI{10}{}$]
& $[\tau_{lb},\tau_{ub}]$ (min) & $[5,60]$ \\

$[\epsilon_{\mathcal{C}},\epsilon_{\mathcal{P}}]$ & $[100,10]$
& $[\epsilon_{r_{h_1}},\epsilon_{r_1},\epsilon_{r_{h_2}},\epsilon_{r_2}]$ (m) & $[30,30,15,5]$ \\

$[\Delta t_g,\delta t_g]$ (s) & $[30,2]$
& $[\Delta t_{CP_1},\Delta t_{CP_2}]$ (s) & $[30,10]$ \\

$\boldsymbol{x}^{\mathrm{rel}}_{i_1}$ (m, m/s) & $[0,\ -\frac{r_{AS}}{2},\ 0,\ 0,\ 0,\ 0]^\top$
& $\boldsymbol{x}^{\mathrm{rel}}_{f_1}$ (m, m/s) & $[\overline{r}_{KOS}\hat{\boldsymbol{e}},\ 0,\ 0,\ 0]^\top$ \\

$\boldsymbol{x}^{\mathrm{rel}}_{f_2}$ (m, m/s) & $[\hat{\boldsymbol{e}},\ 0,\ 0,\ 0]^\top$
& & \\

\bottomrule
\end{tabular}
\label{trajpoints}
\end{table*}

All simulations were conducted on a MacBook Air (M2, 2022) with an 8-core CPU and 8 GB unified memory.

\subsubsection{Initial Reference Generation}
This section presents the results of the initial reference generation process for $N = 100$ uniformly distributed docking port locations. The docking directions are sampled using a Fibonacci sphere distribution \cite{keinert2015spherical} around the \ac{KOS}, ensuring even coverage over the unit sphere.

\begin{figure}[tb]
    \centering
    \begin{subfigure}[b]{0.48\linewidth}
        \centering
        \includegraphics[width=\linewidth]{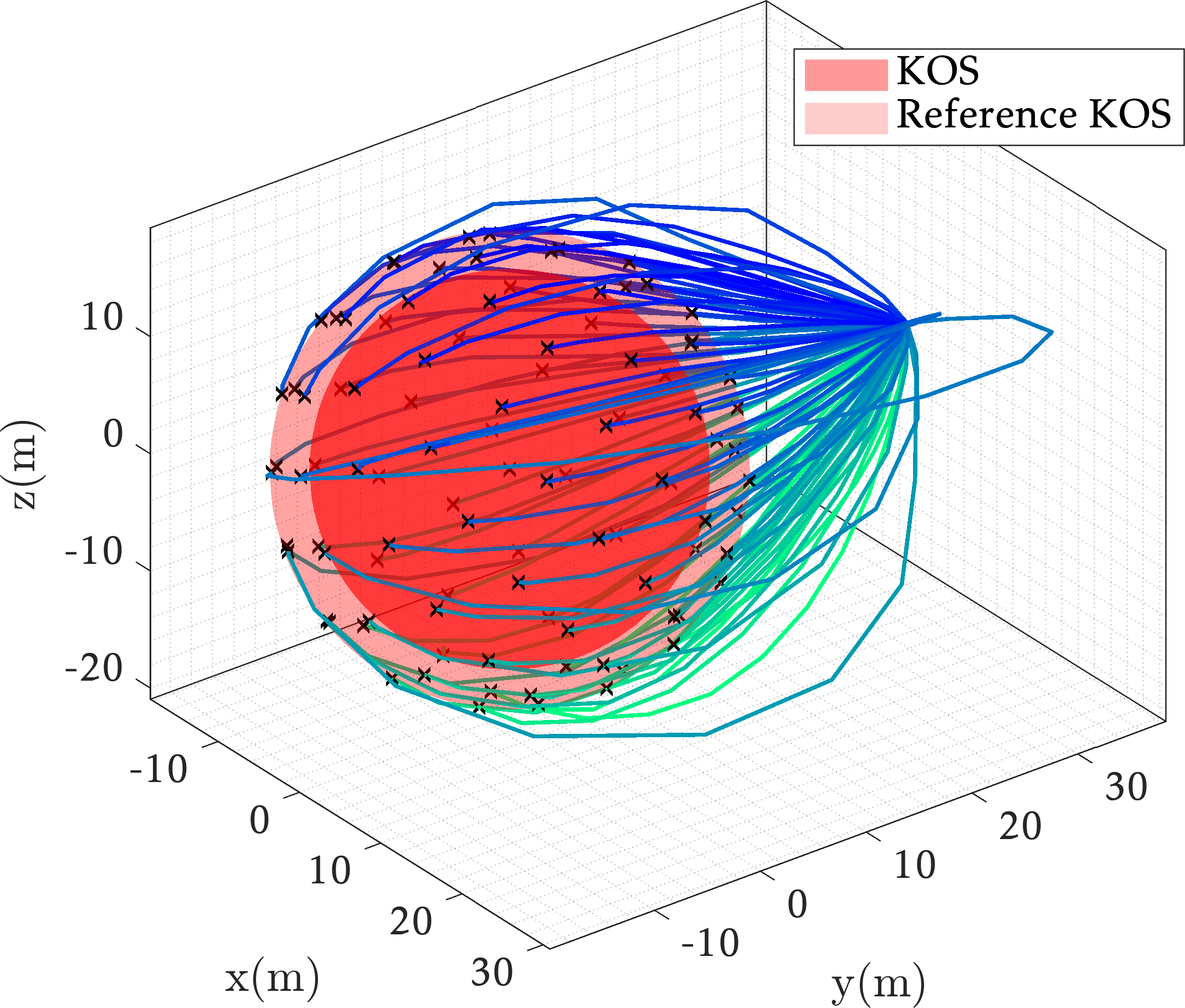}
        \caption{Fly-around reference trajectories}
        \label{fig:flyaroundref}
    \end{subfigure}
    \hfill
    \begin{subfigure}[b]{0.48\linewidth}
        \centering
        \includegraphics[width=\linewidth]{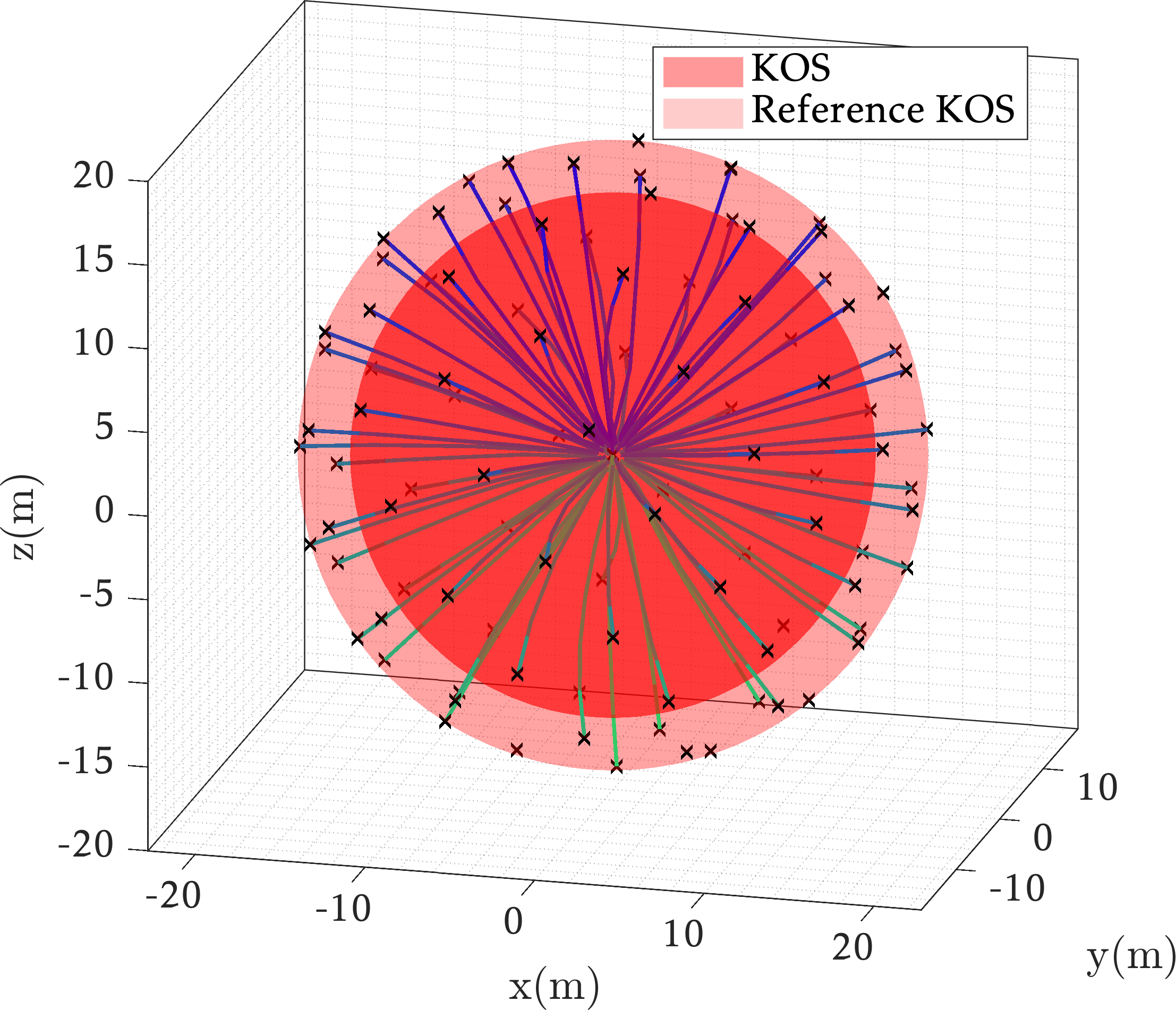}
        \caption{final approach reference trajectories}
        \label{fig:dockingRef}
    \end{subfigure}
    \caption{Reference trajectories for (a) fly-around and (b) final approach.}
    \label{fig:refTrajectories}
\end{figure}

\begin{figure}[tb]
    \centering
    \includegraphics[width=1\linewidth]{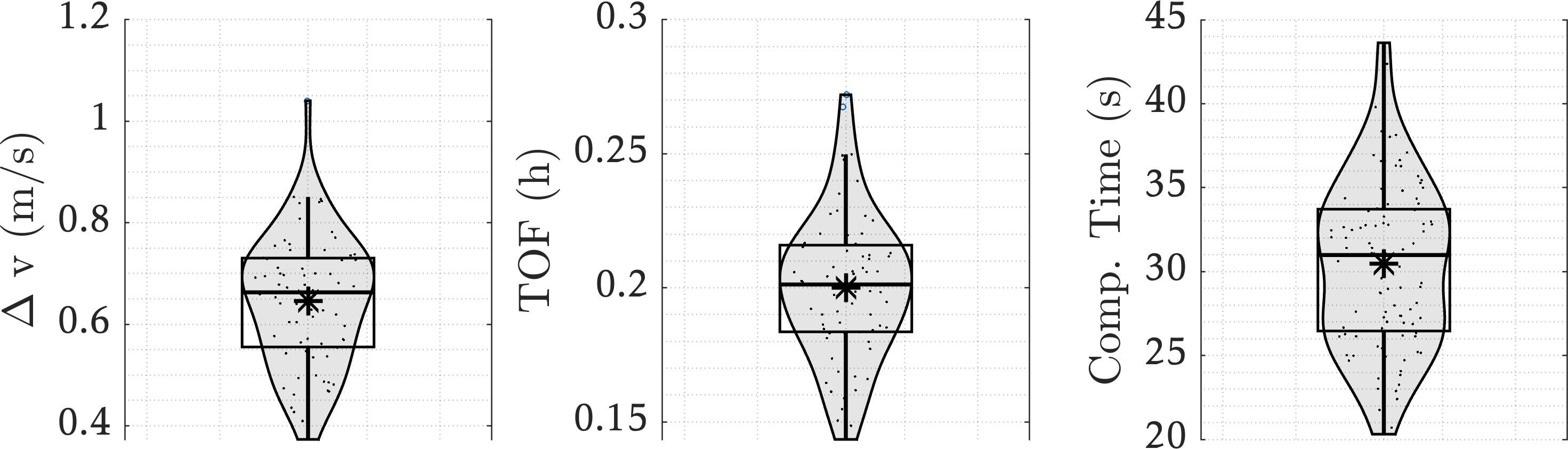}
    \caption{Distribution of computational time, time of flight and fuel consumption for the reference trajectories}
    \label{fig:TOFdvref}
     
\end{figure}

The corresponding reference trajectories are illustrated in Figure~\ref{fig:refTrajectories}, with total time-of-flight (TOF) and fuel cost (\( \Delta v \)) distributions shown in Figure~\ref{fig:TOFdvref}. The median fuel consumption is \SI{0.6579}{\meter\per\second}, with an \ac{IQR} of \SI{0.1600}{\meter\per\second}. The median TOF is \SI{12.07}{\minute}, with an \ac{IQR} of \SI{1.7723}{\minute}. The median computational time is \SI{30.9}{\second}, with an \ac{IQR} of \SI{7.23}{\second} and standard deviation of $\SI{5.01}{\second}$. These results demonstrate that the reference generation strategy reliably and efficiently produces feasible trajectories across a range of docking axes—an essential capability for enabling onboard reference regeneration.

\subsubsection{Convex-tracking guidance}\label{matlabsims}
In this section, the performance of the convex tracking scheme is analyzed under thrust and state uncertainties and missed thrust events via a 100-sample \ac{MC} simulation. First, a reference is generated for a fixed docking axis at
$\hat{\boldsymbol{e}} = [-\sin(\pi/4), -\sin(\pi/4)]^T$, 
which yields a nominal \ac{TOF} of \SI{19.66}{\minute} (Fly around: \SI{14.66}{\minute}, Final Approach: \SI{4.5}{\minute}) and a nominal \(\Delta v\) cost of \SI{1.1740}{\meter\per\second}.

\begin{figure*}[tbp]
    \centering
    \includegraphics[width=\linewidth]{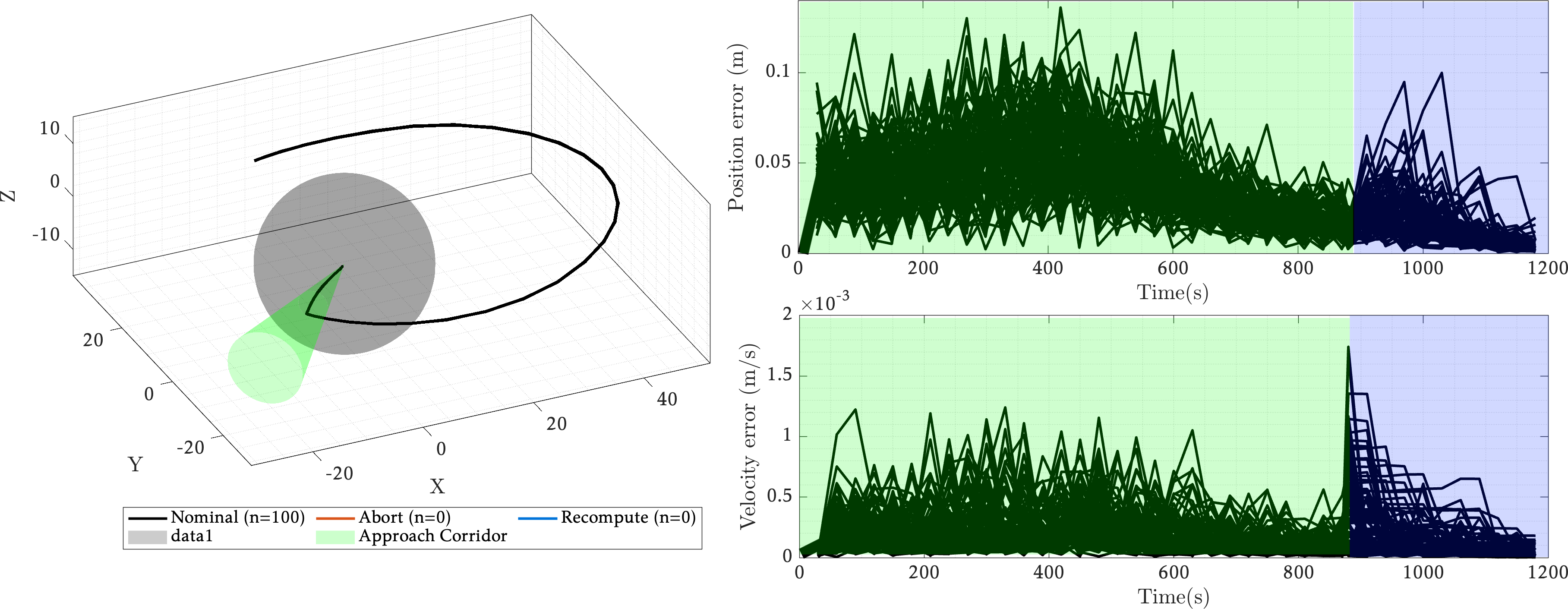}
    \caption{Low Error Case: Propagation of the trajectory and state errors over time. (Green Background: Fly around, Blue Background: Final Approach) }
    \label{fig:MCprop1}
\end{figure*}

\begin{figure*}[tbp]
    \centering
    \includegraphics[width=\linewidth]{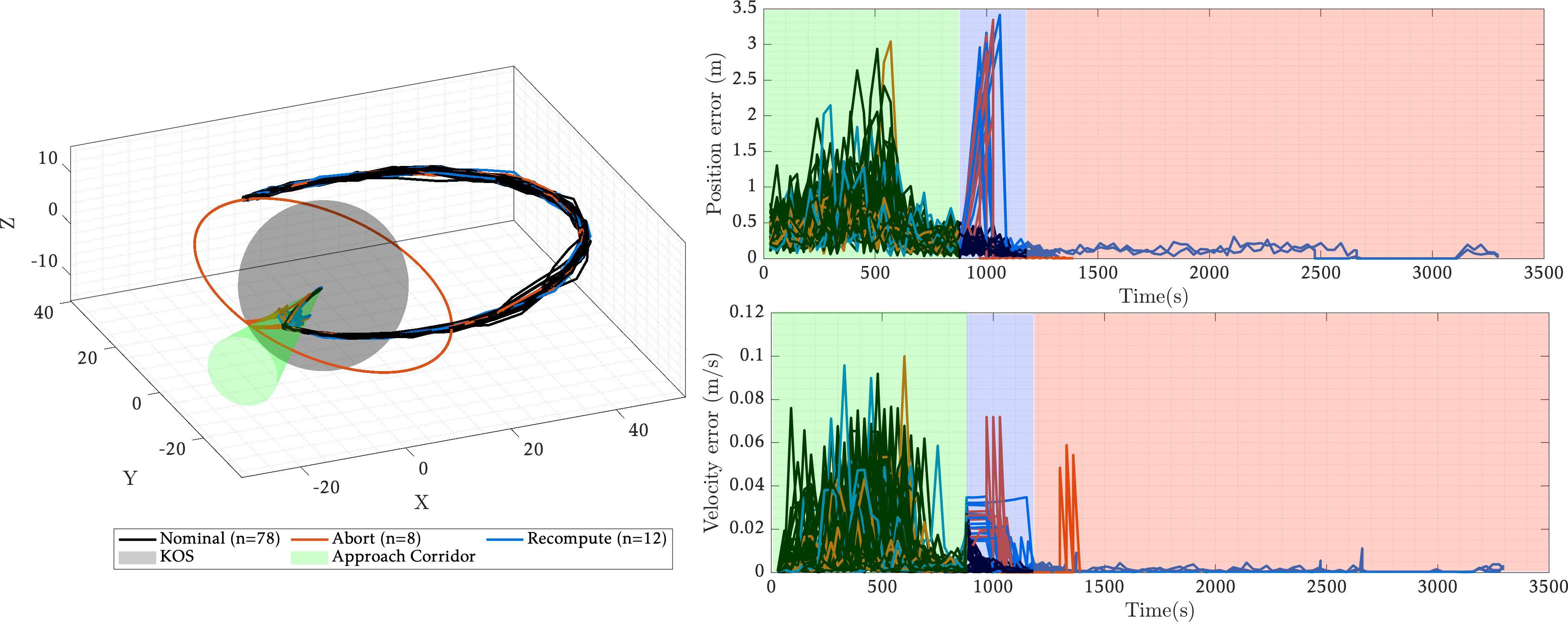}
    \caption{High Error Case: Propagation of the trajectory and state errors over time.  (Green Background: Fly around, Blue Background: Final Approach, Red Background: TOF extensions due to abort and recompute) }
    \label{fig:MCprop2}
\end{figure*}

\begin{figure}[tbp]

    \centering
    \includegraphics[width=1\linewidth]{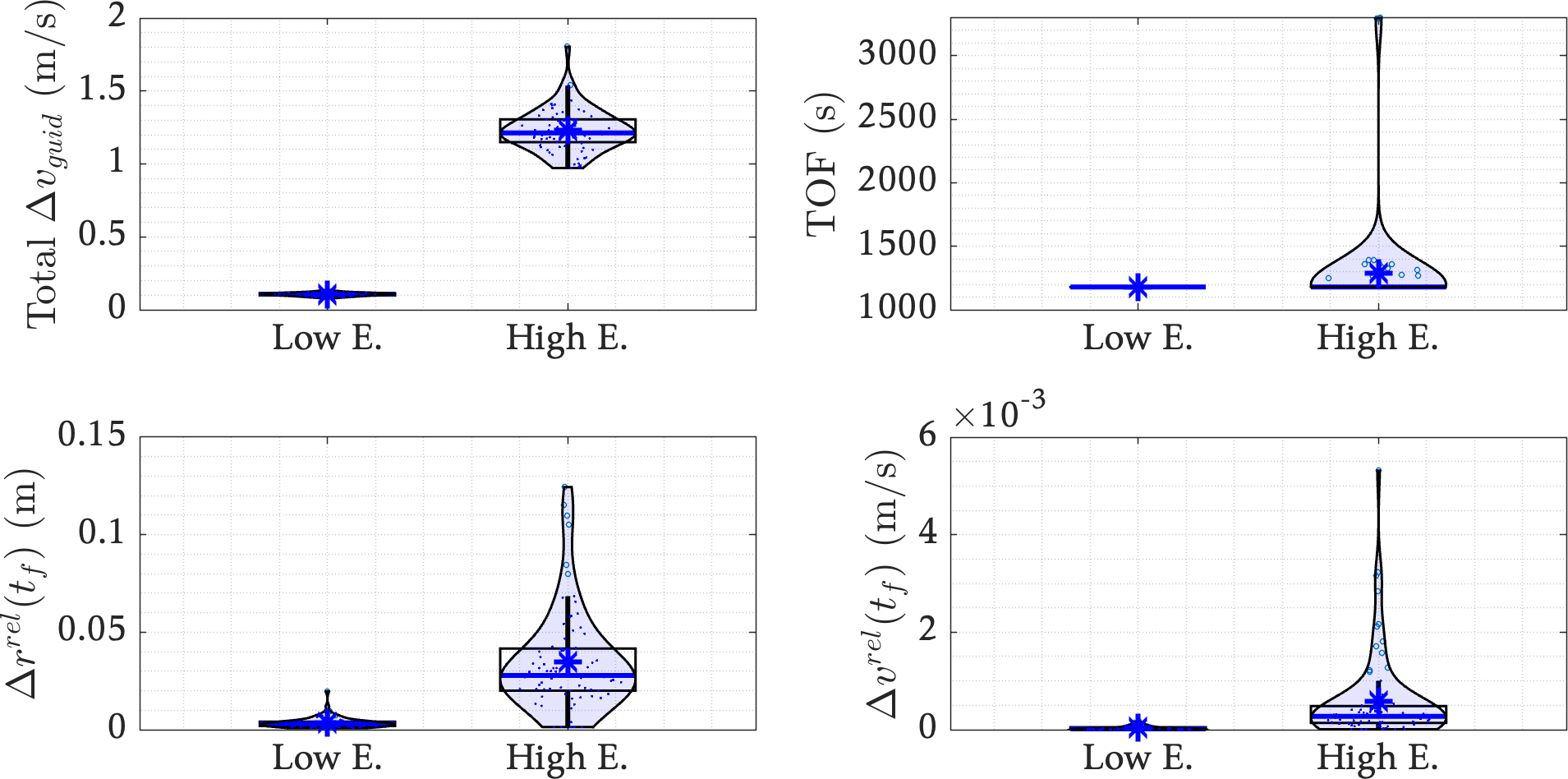}
    \caption{Monte-Carlo Simulation Results from the software testbed.}
    \label{fig:MCdata}
     
\end{figure}

Here the \(3\sigma\) position uncertainty is modelled as a function of $r^{rel}$ as ${\sigma}_{r^{rel}} = \Delta r/3 \left( 0.02 + \left(1 - 0.02\right)\frac{\parallel \boldsymbol{r}^{rel}\parallel}{r_{\text{AS}}} \right)$ Then, ${\sigma}_{v^{rel}}$ is set to be $0.001{\sigma}_{r^{rel}}$. Two different error levels are chosen for the study, as defined in Table \ref{MCerrors}.  The results of the 100-sample \ac{MC} simulation are shown in Figures~\ref{fig:MCprop1}, \ref{fig:MCprop2}  and~\ref{fig:MCdata}. 

\begin{table}[tb]
\centering
\caption{Error levels used in the Monte Carlo simulations of convex guidance}
\begin{tabular}{lccccc} \toprule
Error Level & $\Delta r$ & $\sigma_{\Delta v}$ & $\sigma_\beta$ & $\sigma_\alpha$ & $p_{\text{misthrust}}$ \\
            & (m)        & (m/s)               & (deg)          & (deg)           & --                      \\ \midrule
Low Error   & 0.1        & 0.1                 & 0.5            & 0.5             & 5\% \\
High Error  & 1.0       & 0.2                 & 1.0            & 1.0             & 10\% \\ \bottomrule
\end{tabular}
\label{MCerrors}
     
\end{table}

Figures~\ref{fig:MCprop1} and~\ref{fig:MCprop2} show representative trajectories and the evolution of relative position and velocity errors for low- and high-error Monte Carlo cases. 
No guidance recomputations were required in the low-error regime. 
In contrast, several high-error cases trigger mid-course replanning (blue) and, in the most adverse realisations, aborts (red). 
Replanning is initiated when the \emph{buffered} safety envelopes are violated---specifically the buffered \ac{KOS}, buffered approach corridor, or buffered plume-impingement constraints---or when the vehicle deviates substantially from the reference trajectory. 
These recomputations increase time-of-flight and propellant consumption relative to the low-error case, but still achieve successful docking. 
Aborts are triggered only when the \emph{true} (non-buffered) safety constraints are violated (i.e., \ac{KOS}, approach corridor, or plume impingement), after which the servicer executes a retreat to a passively safe orbit under guidance. 
Notably, several cases that would have resulted in abort without a replanning capability are instead recovered via mid-course recomputation, thereby preserving mission success across a wider range of off-nominal conditions.

Figure~\ref{fig:MCdata} and Table~\ref{tableerror} summarize the corresponding fuel consumption, total time-of-flight (\ac{TOF}), and terminal relative position and velocity errors across all MC realizations. 
As expected, the high-error scenarios exhibit increased fuel usage and longer \ac{TOF}, with both distributions displaying heavy upper tails driven by mid-course replanning and retreats to the passively safe orbit. 
In the low-error results, all trajectories follow the nominal guidance profile closely resulting in identical \ac{TOF} values and tightly clustered propellant consumption. 
Under high error, the mean \ac{TOF} increases to $1271.93$~s with a standard deviation of $363.21$~s, while the $99$th percentile reaches $3296.50$~s due to long-duration recovery manoeuvres following recomputations. 
Similarly, the total $\Delta v$ increases from $0.11\pm0.01$~m/s in the low-error case to $1.23\pm0.14$~m/s under high error. 

Despite these increases, terminal tracking accuracy remains high in both error cases. 
For the high-error scenarios, the median terminal position error is $31.26$~mm and the median terminal velocity error is $0.5243$~mm/s.
Although the upper percentiles of both metrics increase due to severe disturbance realisations, the majority of cases remain tightly bound.
It was also noted that a single convex tracking step of duration $\Delta t_g = 30$~s requires only $38.43 \pm 12.8$~ms of computational time, confirming the suitability of the approach for real-time onboard implementation.

\begin{table}[tbp]
\centering
\footnotesize
\caption{Summary statistics from Monte Carlo simulations. Quartiles column includes 99th percentile ($P_{99}$) as the fourth entry.}
\begin{tabular}{lrrr} \toprule 
\textbf{Metric} & $\mu$ & $\sigma$ & $[Q_1,\ Q_2,\ Q_3,\ P_{99}]$ \\ \midrule

\multicolumn{4}{c}{\textbf{Low Error}} \\
 Total TOF (s)                 & 1179.82 & 0.00 &  [1179.82, 1179.82, 1179.82,1179.82] \\
Total $\Delta v$ (m/s)        & 0.11 & 0.01 & [0.10, 0.10, 0.11,0.13]  \\
$\Delta r^{rel}$ ($t_f$) (mm)  & 3.67 & 3.32 & [2.23, 2.88, 3.96,21.05] \\
 $\Delta v^{rel}$ ($t_f$) (mm/s)& 0.0335 & 0.0388 & [0.0079, 0.0181, 0.0426,0.18]  \\
\midrule

\multicolumn{4}{c}{\textbf{High Error}} \\
 Total TOF (s)                 & 1271.93 & 363.21 & [1179.82, 1179.82, 1179.82,3296.50] \\
Total $\Delta v$ (m/s)        & 1.23 & 0.14 & [1.14, 1.21, 1.30,1.67]  \\
$\Delta r^{rel}$ ($t_f$) (mm)  & 31.26 & 24.17 & [19.18, 25.53, 38.64,119.85] \\
 $\Delta v^{rel}$ ($t_f$) (mm/s)& 0.5243 & 0.8034 & [0.1195, 0.2747, 0.4869,4.27]  \\
\bottomrule 
\end{tabular}
\label{tableerror2}
  
\end{table}

\begin{figure*}[tbp]
    \centering
    \begin{subfigure}[t]{0.567\linewidth}
        \centering
        \includegraphics[width=\linewidth]{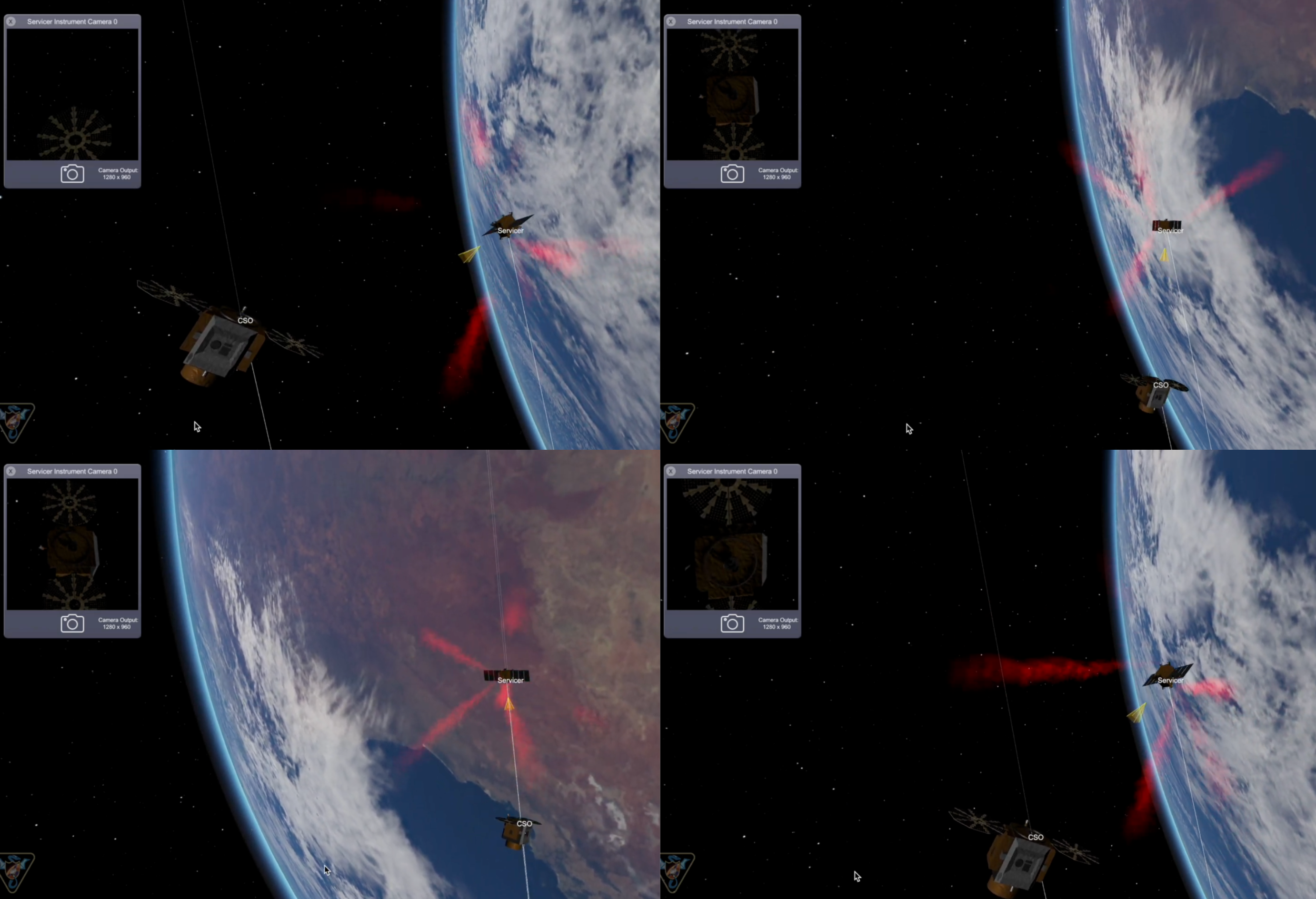}
        \caption{Fly-around docking-port localisation.}
    \end{subfigure}\hfill
    \begin{subfigure}[t]{0.43\linewidth}
        \centering
        \includegraphics[width=\linewidth]{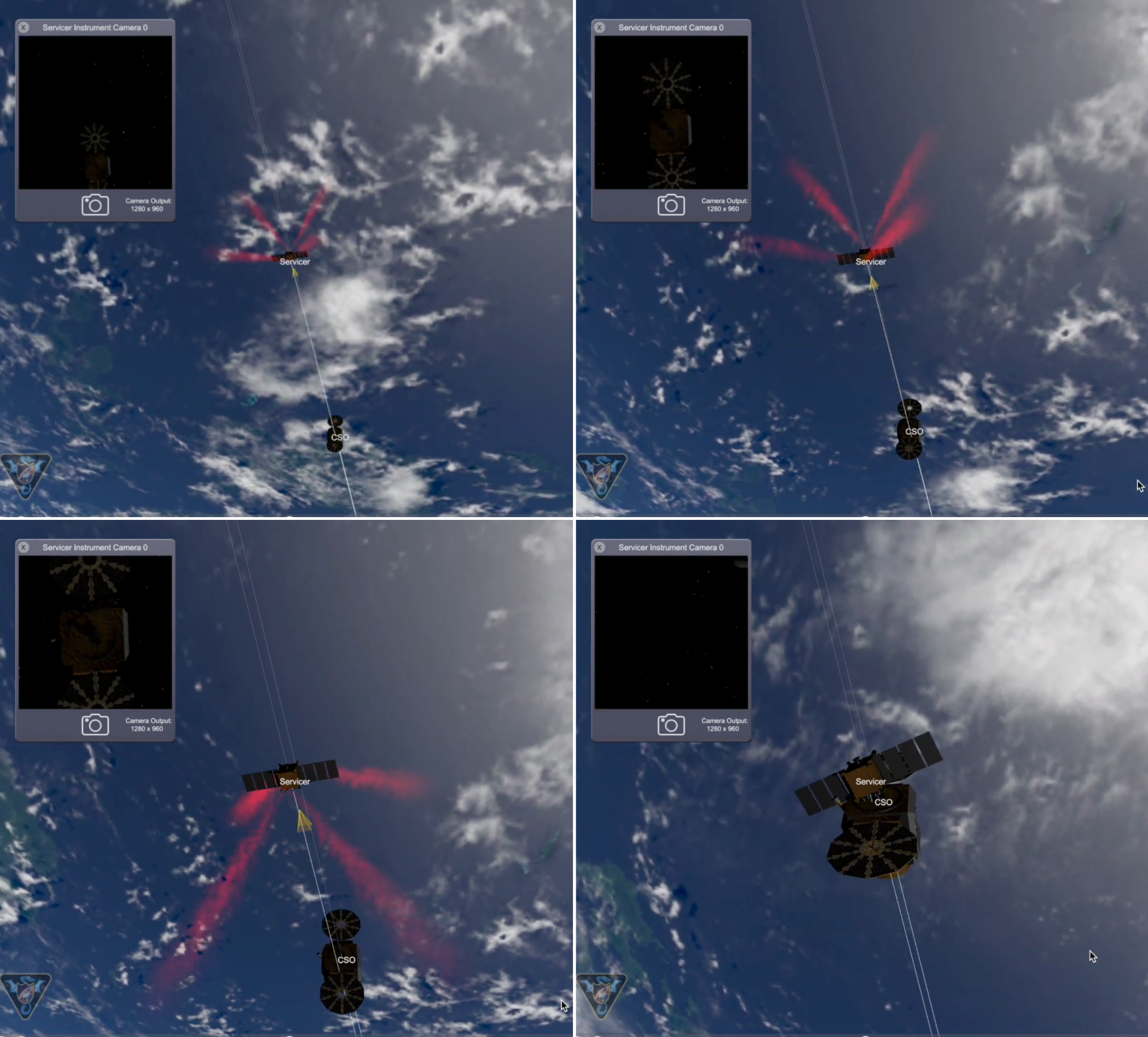}
        \caption{Final approach trajectory.}
    \end{subfigure}
    \caption{Basilisk simulations using on-board convex guidance: (a) fly-around docking-port localisation and (b) final approach trajectory.}
    \label{fig:basilisk_flyaround_and_approach}
\end{figure*}

Lastly, Figure~\ref{fig:basilisk_flyaround_and_approach} shows representative snapshots from the Basilisk simulation during the fly-around and final approach phases. 
During the fly-around manoeuvre, the servicer maintains a safe relative orbit while actively reorienting to localize the target docking interface. 
Following successful docking-port localization, the vehicle transitions into the final approach phase, executing a constrained closing trajectory while maintaining continuous target pointing and approach-corridor compliance.

\subsection{Hardware Testbed Results}
This section evaluates the performance of the proposed guidance framework using planar air-bearing hardware experiments under both nominal and off-nominal conditions. 

A set of $10$ nominal trials is first conducted to assess repeatability and baseline performance. For these cases, both time-averaged tracking errors and terminal relative position and velocity errors at $t_f$ are computed and analyzed. A small series of off-nominal test cases is then considered to examine robustness and safety-handling behaviour under representative disturbances and failure cases. These include:
\begin{enumerate}
    \item A high thruster jitter and elevated actuation noise leading to an abort ($\times 2$);
      \item A high thruster jitter and elevated actuation noise with recovery ($\times 2$);
          \item A missed-thrust event followed by recovery ($\times 2$);

    \item A \ac{KOS} violation that result in an abort;
    
    \item recovery from an unexpected external perturbation, represented by a small physical push. 

\end{enumerate}

Together, these experiments demonstrate the ability of the guidance architecture to maintain accuracy under nominal conditions, recover from moderate disturbances, and enforce safety through abort logic when constraint violations occur.

\subsubsection{Nominal Cases}

\begin{figure*}[tbp]
    \centering
    \includegraphics[width=\linewidth]{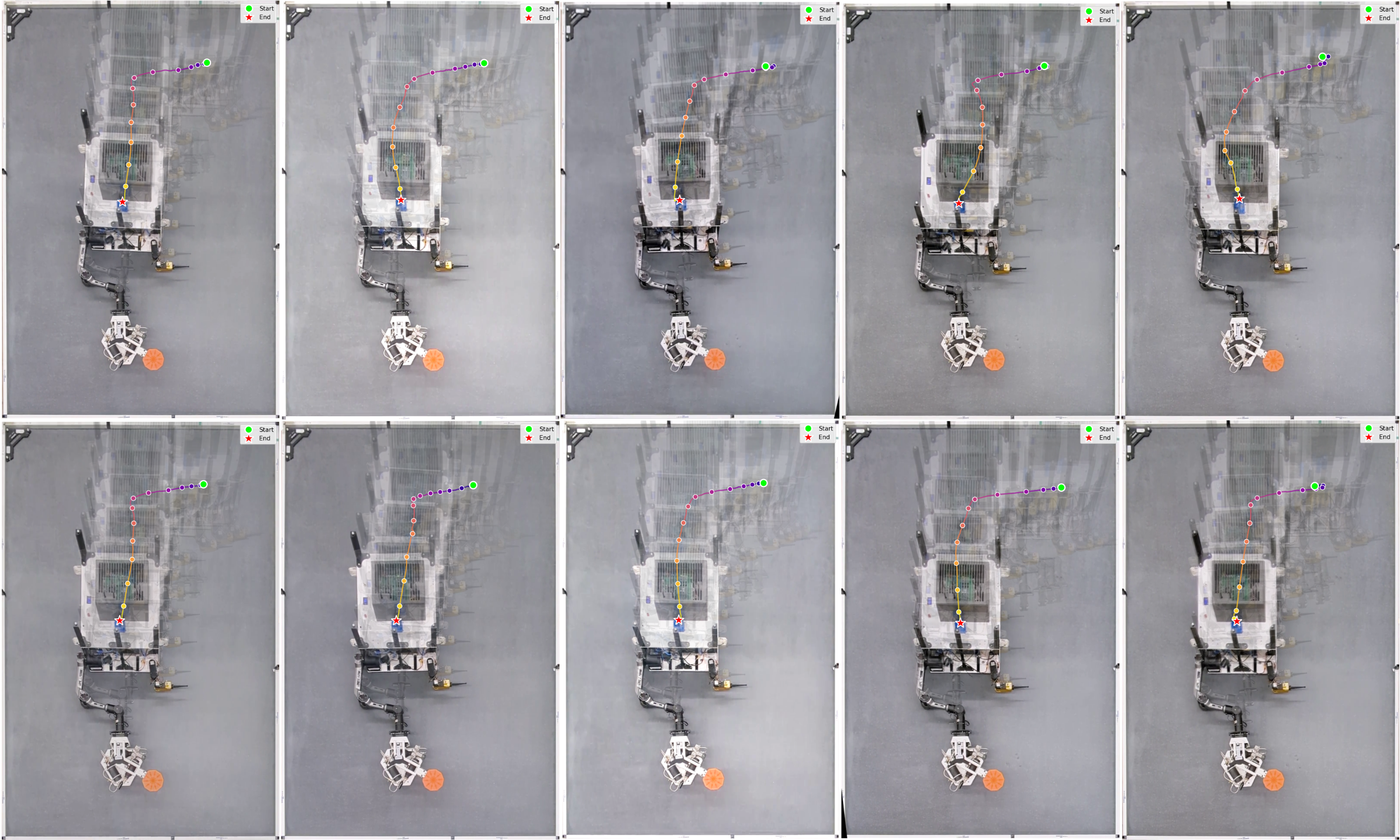}
    \caption{Timelapse of all 10 nominal MC trials. Green denotes the start while red point denotes the end. The keypoints shown along each trajectory are distributed evenly in time, every $20$ seconds. }
    \label{fig:nominal}
\end{figure*}

\begin{figure}[tbp]
    \centering
    \includegraphics[width=\linewidth]{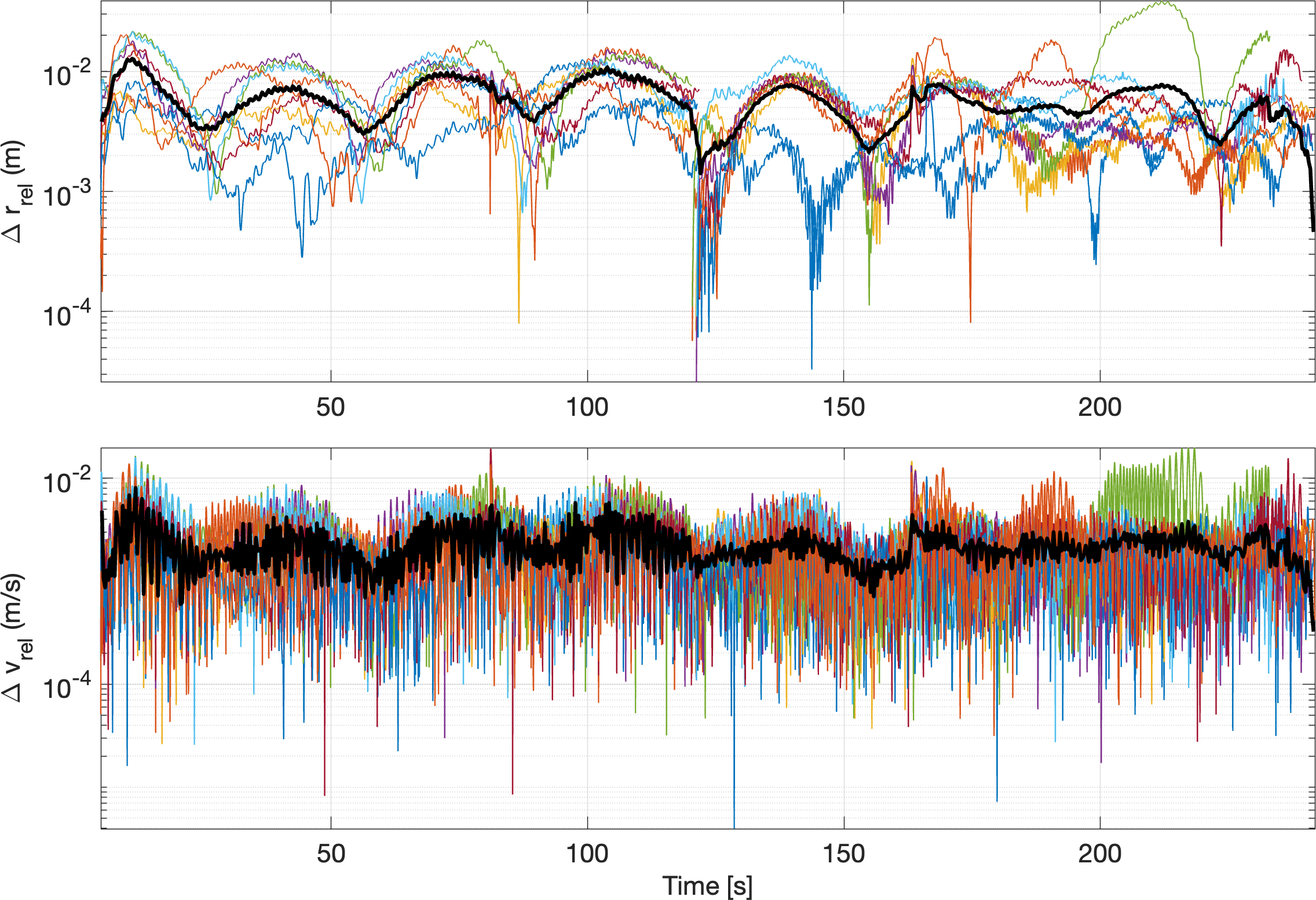}
    \caption{Position and velocity tracking errors with respect to the reference trajectory for all nominal hardware testbed CORTEX simulation trials during the docking phase.}
    \label{fig:drdvExp}
\end{figure}

The ten nominal test cases are shown in Fig.~\ref{fig:nominal}. For each trajectory, the starting point is indicated by a green dot and the endpoint by a red star. Keypoints are overlaid along each trajectory at uniform temporal intervals, with one marker every 20~s. Figure~\ref{fig:drdvExp} illustrates the evolution of relative position and velocity errors (with respect to the reference) over time, and Table~\ref{terrorhard} reports the corresponding terminal error statistics across all trials.

As summarized in Table~\ref{tableerror}, the nominal-case terminal relative position error remains on the order of a few millimetres, with a mean value of $3.67$~mm and median below $3$~mm, while terminal relative velocity errors remain below $6.2$~mm/s for most cases. These error magnitudes indicate stable convergence of the closed-loop guidance despite unmodeled disturbances, residual drift, and actuator imperfections inherent to the planar air-bearing environment. The observed terminal error levels also closely match those obtained in the software simulation campaign. 




\begin{table}[h]
\centering
\footnotesize
\caption{Terminal position and velocity tracking error statistics from hardware testbed experiments.}
\label{terrorhard}
\begin{tabular}{lrrr} \toprule 
Metric                          & $\mu$  & $\sigma$ & $[Q_1,\ Q_2,\ Q_3,\ P_{99}]$  \\ \hline
\multicolumn{4}{l}{\textit{Nominal cases}}                                          \\
$\Delta r^{rel}$ ($t_f$) (mm)   & 6.25   & 5.35     & [3.62, 4.80, 5.91,19.77]      \\
$\Delta v^{rel}$ ($t_f$) (mm/s) & 2.35   & 1.68     & [1.32, 1.93, 2.79,6.20]       \\
\multicolumn{4}{l}{\textit{Off-nominal cases}}                                      \\
$\Delta r^{rel}$ ($t_f$) (mm)   & 8.09   & 5.29     & [4.41, 5.22, 12.35,15.92]     \\
$\Delta v^{rel}$ ($t_f$) (mm/s) & 2.2316 & 1.7175   & [0.7382, 2.3107, 3.1240,4.89] \\ 
\bottomrule 
\end{tabular}
\label{tableerror}
  
\end{table}


\subsubsection{Off-nominal cases}

\begin{figure*}[tbp]
    \centering
    \includegraphics[width=0.8\linewidth]{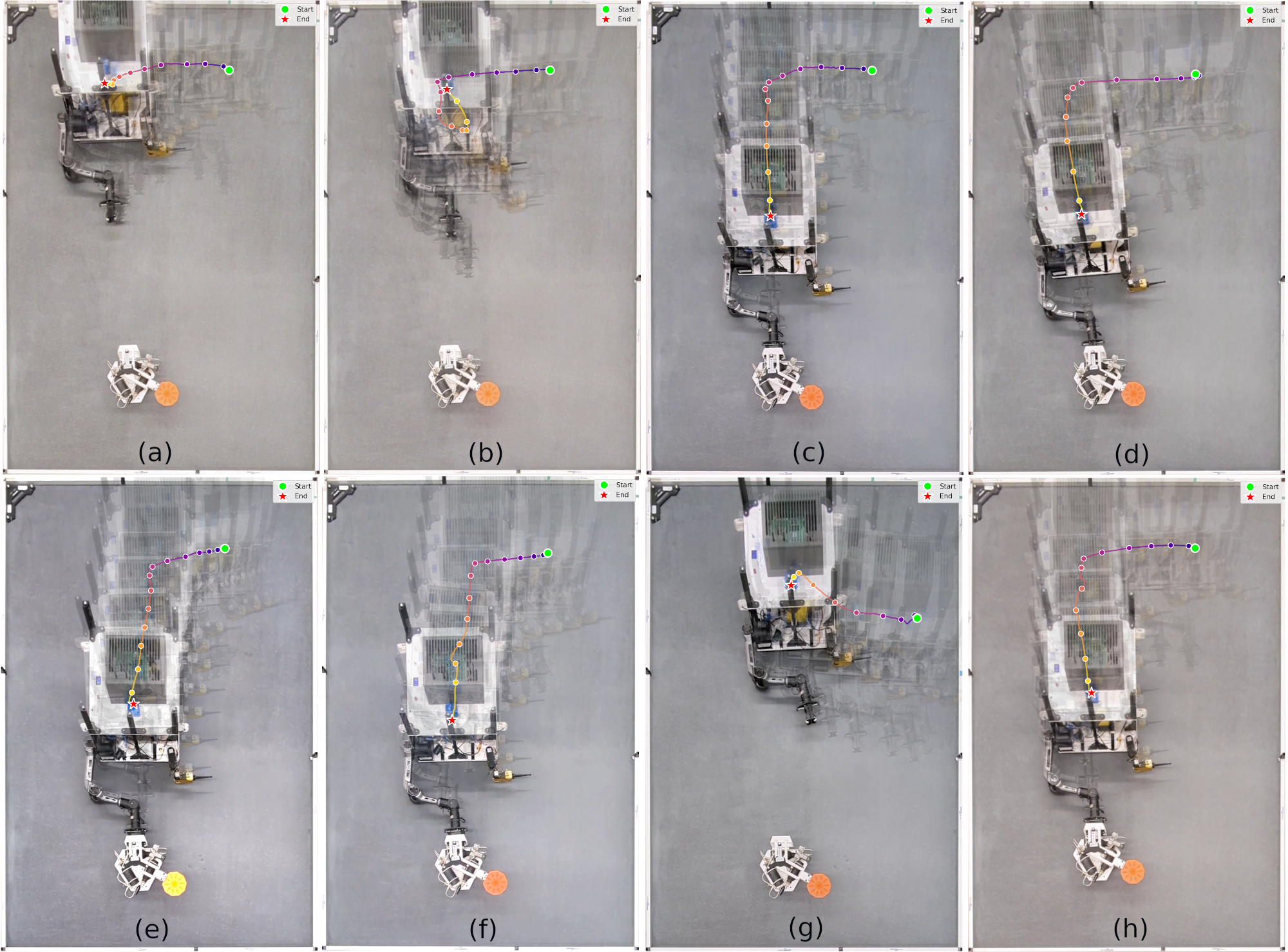}
    \caption{Timelapse of the off-nominal cases. (a-b) high state error triggering abort, (c-d) high state error with recovery, (e-f) missed thrust event with recovery, (g) Abort triggered by KOS violation, (h) recovery from physical perturbation.}
    \label{fig:offnomtimelapse}
\end{figure*}

The set of off-nominal scenarios tested are shown in Fig.~\ref{fig:offnomtimelapse}.  
Figs.~\ref{fig:offnomtimelapse}(a–b) show cases in which Gaussian thrust disturbances and state errors are introduced, leading to immediate aborts. In contrast, Figs.~\ref{fig:offnomtimelapse}(c–d) demonstrate successful recovery from similar state errors, where the guidance law re-establishes a safe approach corridor and resumes convergence. Recovery from a 20 s missed thrust event is illustrated in Figs.~\ref{fig:offnomtimelapse}(e–f), where transient deviations are corrected without violating safety constraints. Fig.~\ref{fig:offnomtimelapse}(g) shows an abort triggered by a keep-out-zone (KOZ) violation, while Fig.~\ref{fig:offnomtimelapse}(h) demonstrates recovery following a deliberate physical perturbation applied to the servicer.

\begin{figure}[tbp]
    \centering
    \includegraphics[width=\linewidth]{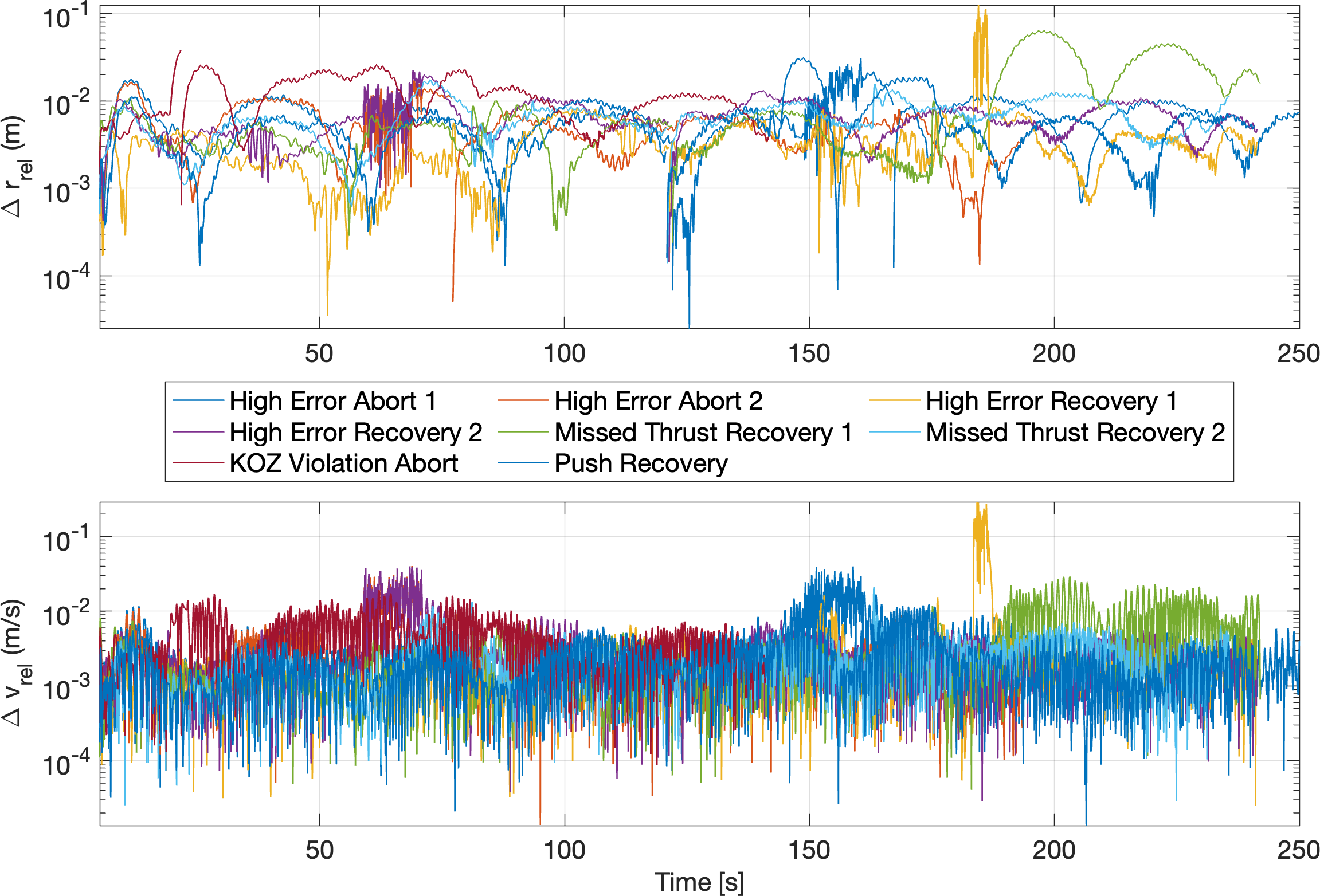}
    \caption{Position and velocity tracking errors with respect to the reference trajectory for all off-nominal hardware testbed CORTEX simulation trials during the docking phase.}
    \label{fig:offnompos}
\end{figure}

The corresponding position and velocity tracking errors for all off-nominal trials are shown in Fig.~\ref{fig:offnompos}. As expected, off-nominal conditions induce larger transient errors than nominal cases; however, in recovery scenarios, these errors remain bounded and decay as tracking continues. Terminal error statistics for the off-nominal cases are reported in Table~\ref{terrorhard}. Despite the presence of disturbances and execution faults, the mean terminal relative position and velocity errors remain on the order of a few millimetres and millimetres per second, respectively, for all recovered trials.

Importantly, no unsafe behaviour is observed: all trajectories either recover while respecting the imposed safety constraints or are terminated via an abort before entering unsafe regions. These results demonstrate that CORTEX provides both robustness and verifiable safety under realistic hardware-induced disturbances and execution uncertainties.

\section{Conclusion}

This paper presented CORTEX, an autonomous, convex-optimization-based framework for close-range rendezvous trajectory design and guidance. The framework integrates a hierarchical reference generation strategy that couples eclipse-aware phase scheduling with convex trajectory optimization for fly-around and docking phases, a single-iteration convex adaptive tracking controller for impulsive proximity operations, and operational autonomy logic including reference regeneration triggers and abort-to-safe-orbit capability under off-nominal conditions. Relative navigation is provided by a deep-learning perception pipeline based on YOLO keypoint detection fused with an Extended Kalman Filter.

CORTEX was validated through a two-tier campaign spanning high-fidelity software simulation and hardware-in-the-loop experiments. In the software testbed, built on the Basilisk astrodynamics framework, a 100-sample Monte Carlo study was conducted under two error regimes incorporating initial state uncertainty, thrust magnitude, and pointing errors, and missed-thrust events. Under low-error conditions, all runs completed nominally with terminal position errors of $3.67 \pm 3.32$ mm and velocity errors of $0.034 \pm 0.039$ mm/s. Under high-error conditions, the framework demonstrated its adaptive capabilities, where 12 of 100 cases triggered mid-course reference regeneration, and 8 triggered abort-to-safe-orbit manoeuvres, while the remaining cases completed nominally. Even under these adverse conditions, terminal docking accuracy for successfully completed runs remained at $31.26$ mm (median position) and $0.52$ mm/s (median velocity). A single convex guidance solve required only $38.43 \pm 12.8$ ms, confirming the framework's suitability for real-time onboard implementation.
Hardware validation was performed on a planar air-bearing testbed using a fully integrated servicer platform with cold-gas thrusters, an onboard computer, and motion-capture-based state feedback. Ten nominal trials with no artificially added errors achieved terminal position errors of $6.25 \pm 5.35$ mm and velocity errors of $2.35 \pm 1.68$ mm/s, closely matching the software simulation results. Eight off-nominal cases are introduced to show the full range of CORTEX's contingency handling, including recovery from artificially elevated actuation noise, missed-thrust events, and physical perturbations, as well as abort execution upon \ac{KOS} violations. In all cases, the framework either recovered successfully while respecting safety constraints or executed a safe retreat before entering unsafe regions, with median terminal position errors of $8.09 \pm 5.29$ mm and median velocity errors of $2.23 \pm 1.72$ mm/s.

Several directions remain for future work. The plume impingement constraint is currently handled with a fixed-orientation assumption; thus, coupling this constraint with real-time attitude control would improve fidelity during the final approach. Integrating non-convex constraints such as the keep-out sphere and plume impingement directly into the convex guidance formulation, potentially through successive convexification within the tracking loop, could reduce reliance on conservative reference margins and improve fuel efficiency. Finally, closing the loop with the onboard vision-based perception pipeline in the hardware testbed, rather than relying on external motion capture, would constitute a key step toward flight-representative demonstration of the full autonomous system.

\section*{Acknowledgements}
This work has been supported by the SmartSat CRC, whose activities are funded by the Australian Government’s CRC Program.
The authors thank Tim Bailey, Will Thorp, and Viorela Ila for their contributions to the perception pipeline, and Jiashu (George) Wu for his support in building and operating the planar air-bearing testbed.

\bibliographystyle{plainnat}  
\bibliography{library}        

\end{document}